%% file: main.tex
\RequirePackage[svgnames]{xcolor}
\RequirePackage[dvipsnames]{xcolor}

\documentclass[twocolumn]{mystyle}

\usepackage[all]{hypcap}
\usepackage[svgnames]{xcolor}
\usepackage[dvipsnames]{xcolor}

\usepackage[numbers]{natbib}
\usepackage{hyperref}

\usepackage{algorithm}
\usepackage{algorithmicx}
\usepackage{algpseudocode}
\usepackage{microtype}
\usepackage{graphicx}
\expandafter\def\csname ver@subfig.sty\endcsname{}
\usepackage{booktabs} 
\usepackage{float}
\usepackage{bigstrut}

\usepackage{amsmath}
\usepackage{amssymb}
\usepackage{mathtools}
\usepackage{amsthm}
\usepackage{mathrsfs}
\usepackage{nicefrac}
\usepackage{dsfont}
\usepackage{enumitem}
\usepackage{subcaption}

\usepackage{graphicx,subfig}
\usepackage{cleveref}
\usepackage{bxcoloremoji}

\usepackage{float}

\usepackage[utf8]{inputenc} %
\usepackage[T1]{fontenc}    %
\usepackage{hyperref}       %
\usepackage{url}            %
\usepackage{booktabs}       %
\usepackage{amsfonts}       %
\usepackage{nicefrac}       %
\usepackage{microtype}      %
\usepackage{graphicx}
\usepackage{subcaption} 
\usepackage{amssymb}
\usepackage{fdsymbol}
\usepackage{wrapfig}
\usepackage{lipsum}
\usepackage{enumitem}
\usepackage{stackengine}
\usepackage[font=small,labelfont=bf]{caption}
\usepackage{color}
\usepackage{adjustbox}

\usepackage{rotating}
\usepackage{makecell}
\usepackage{xspace}
\usepackage{multirow}
\newcommand{\bm}[1]{\textbf{#1}}

\definecolor{hidden-red}{RGB}{205, 44, 36}
\definecolor{hidden-blue}{RGB}{194,232,247}
\definecolor{hidden-orange}{RGB}{243,202,120}
\definecolor{hidden-green}{RGB}{34,139,34}
\definecolor{hidden-pink}{RGB}{255,245,247}
\definecolor{hidden-black}{RGB}{20,68,106}

\input{mycommands}

\usepackage{amsmath}

\title{\emph{From Verbatim to Gist}:\\Distilling Pyramidal Multimodal Memory via Semantic Information Bottleneck for Long-Horizon Video Agents}
\runningtitle{\emph{From Verbatim to Gist}: Distilling Pyramidal Multimodal Memory via Semantic Information Bottleneck for Long-Horizon Video Agents}

\author[1,2,*]{\mbox{Niu Lian}}
\author[*]{\mbox{Yuting Wang}}
\author[2]{\mbox{Hanshu Yao}}
\author[2,\dagger]{\mbox{Jinpeng Wang}}
\author[2]{\mbox{Bin Chen}}
\author[2,3]{\mbox{Yaowei Wang}}
\author[2]{\mbox{Min Zhang}}
\author[1]{\mbox{Shu-Tao Xia}}
\affil[1]{\mbox{Tsinghua Shenzhen International Graduate School, Tsinghua University}}
\affil[2]{\mbox{Harbin Institute of Technology, Shenzhen}}
\affil[3]{\mbox{Peng Cheng Laboratory}}

\authornote[*]{\mbox{Equal contribution}}
\authornote[\dagger]{\mbox{Corresponding author.}}

\begin{document}
\input{sections/abstract}

\maketitle
\footnotetext[1]{Accepted by ACL 2026 Main Conference.}

\input{sections/introduction}
\input{sections/relatedwork}
\input{sections/method}
\input{sections/experiments}
\input{sections/conclusion}
\bibliography{main}

\clearpage
\appendix
\input{appendix_sections/appendix}

\end{document}

%% file: mycommands.tex
\definecolor{ForestGreen}{RGB}{34,139,34}

\renewcommand{\paragraph}[1]{\medskip\noindent\textbf{#1.~}}

\newcommand{\github}{\raisebox{-1.5pt}{\includegraphics[height=1.05em]{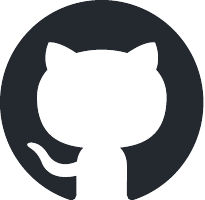}}}

\newcommand{\modelname}{{\scshape MM-Mem}\xspace}

%% file: sections/abstract.tex
\begin{abstract}
While multimodal large language models have demonstrated impressive short-term reasoning, they struggle with long-horizon video understanding due to limited context windows and static memory mechanisms that fail to mirror human cognitive efficiency. Existing paradigms typically fall into two extremes: vision-centric methods that incur high latency and redundancy through dense visual accumulation, or text-centric approaches that suffer from detail loss and hallucination via aggressive captioning. To bridge this gap, we propose \modelname{}, a pyramidal multimodal memory architecture grounded in \emph{Fuzzy-Trace Theory}. \modelname{} structures memory hierarchically into a \emph{Sensory Buffer}, \emph{Episodic Stream}, and \emph{Symbolic Schema}, enabling the progressive distillation of fine-grained perceptual traces (\emph{verbatim}) into high-level semantic schemas (\emph{gist}). 
Furthermore, to govern the dynamic construction of memory, we derive a Semantic Information Bottleneck objective and introduce SIB-GRPO to optimize the trade-off between memory compression and task-relevant information retention. 
In inference, we design an entropy-driven top-down memory retrieval strategy. 
Extensive experiments across 4 benchmarks confirm that \modelname{} achieves state-of-the-art performance on both offline and streaming tasks, demonstrating robust generalization and validating the effectiveness of cognition-inspired memory organization. 

\vspace{2mm}
\vspace{5mm}

\coloremojicode{1F4C5} \textbf{Date}: April 19, 2026

\github{} \textbf{Code Repository}: \href{https://github.com/EliSpectre/MM-Mem}{https://github.com/EliSpectre/MM-Mem}

\coloremojicode{1F4E7} \textbf{Contact}: \href{mailto:220110904@stu.hit.edu.cn}{220110904@stu.hit.edu.cn} (Niu Lian), \href{mailto:wangjp26@gmail.com}{wangjp26@gmail.com} (Jinpeng Wang)

\end{abstract}

%% file: sections/introduction.tex
\section{Introduction}
\label{sec:intro}

\begin{figure}[t]
  \centering
  \includegraphics[width=\linewidth]{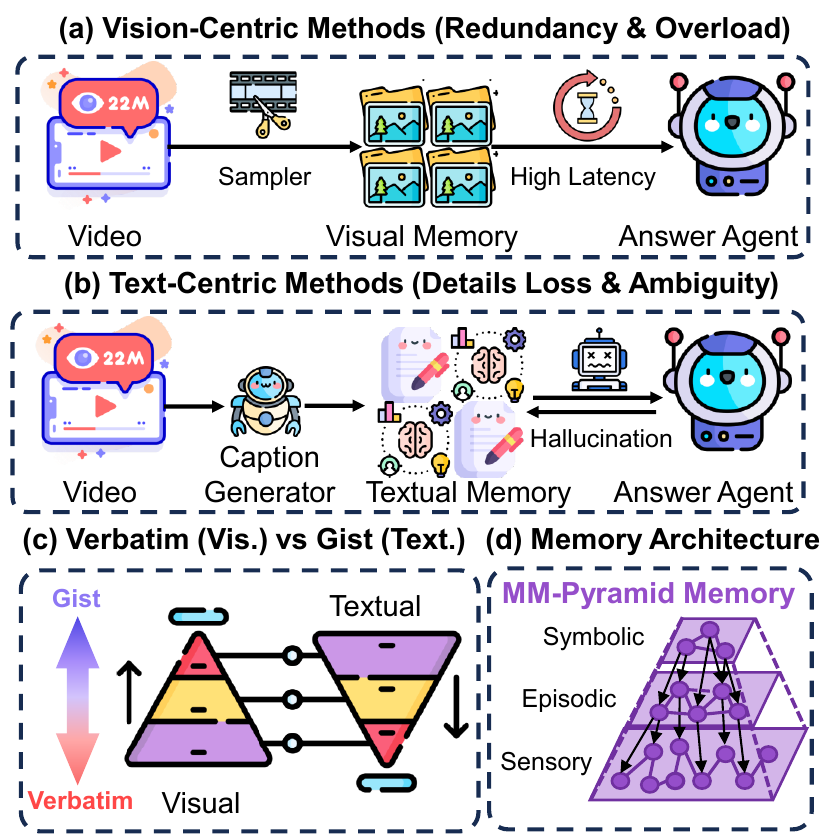}
  \caption{Existing memory paradigms (a-b), inspiration (c), and our insight (d). 
(a) Vision-centric methods incur redundancy and high latency due to dense visual memories.
(b) Text-centric methods suffer from information loss during captioning, leading to hallucination and ambiguity.
(c) The natural complementarity between vision and text neatly aligns with the distinction between verbatim and gist traces in \emph{Fuzzy-Trace Theory}.
(d) Our \modelname{} is a bottom-up multimodal memory pyramid from sensory buffer to symbolic schema.
}
  \label{fig: intro}
\end{figure}

To transition from passive observers to truly autonomous agents, intelligent systems require a critical cognitive shift toward long-term and persistent memory \cite{memory1}, enabling them to move beyond ``here-and-now'' perception and interpret continuous, unbounded streams of multimodal information \cite{memory2}. While recent Multimodal Large Language Models (MLLMs) \cite{llava, blip, blip2} have enhanced short-term perceptual modeling \cite{mmagent1}, they lack the efficient memory mechanisms found in human cognition. Specifically, \textbf{Fuzzy-Trace Theory} (\textbf{FTT}) \cite{FTT}, a well-established cognitive model, hypothesizes that human memory is not a singular recording but consists of two parallel traces: a \textit{gist trace} that captures abstract semantic meaning and a \textit{verbatim trace} that preserves fine-grained perceptual details, allowing to retain specific visual evidence when necessary while efficiently managing long-term semantic context without cognitive overload.

However, existing MLLM-based agents typically fail to strike this biological balance, generally falling into one of two extremes. 
\textbf{Vision-centric paradigms}, such as LongVA \cite{longva} or VideoRAG \cite{VideoRAG}, attempt to continuously accumulate visual memories (Fig.~\ref{fig: intro}(a)). While aiming for fidelity, these designs often introduce substantial redundancy due to dense frame sampling. Furthermore, traditional MLLMs \cite{qwen3vl} utilized in these systems often emphasize low-level visual details while overlooking high-level semantic attributes \cite{M3-Agent}, making it difficult to preserve long-term temporal dependencies.
Conversely, \textbf{text-centric paradigms} \cite{Vgent} convert raw videos into structured textual memories (e.g., knowledge graphs) for efficiency (Fig.~\ref{fig: intro}(b)). Yet, this conversion acts as a lossy compression that discards critical visual cues, leading to ambiguity and hallucinations. 
Moreover, unlike human memory which is highly dynamic, most existing systems remain static. While dynamic memory management is studied in Large Language Models (LLMs) \cite{Memorybank}, it is under-explored in multimodal settings. Even recent attempts like A-Mem \cite{A-mem} remain text-centric, lacking multimodal grounding for long-horizon reasoning.

To bridge this gap, we propose \textbf{\modelname{}}, a novel hierarchical pyramidal multimodal memory architecture inspired directly by the principles of FTT. As illustrated in Fig.~\ref{fig: intro}(c) and (d), \modelname{} is inspired by the complementarity between visual and textual modalities and the distinction between verbatim and gist traces in FTT. Importantly, this connection is realized through cross-modal fusion rather than a rigid one-to-one layer mapping: visual representations predominantly preserve verbatim perceptual evidence, while textual representations mainly encode gist-level semantics. Built upon this principle, \modelname{} spans from perception to cognition across three layers: a \textit{Sensory Buffer} for fine-grained visual evidence, an \textit{Episodic Stream} for event-level summaries, and a \textit{Symbolic Schema} for high-level semantic abstraction. This bottom-up construction progressively transforms perceptual signals into cognitive knowledge.

Crucially, to coordinate transitions across memory layers, we establish a dual mechanism for adaptive construction and efficient retrieval. For memory construction, inspired by the Information Bottleneck theory \cite{Bottleneck}, we derive a principled objective that retains maximal semantic content under a limited budget. Building on this formulation, we propose \textbf{SIB-GRPO} (Semantic-Information Bottleneck GRPO) to balance semantic preservation against redundancy reduction. For retrieval, we further introduce an \emph{entropy-driven} top-down strategy: the agent starts from the abstract Symbolic Schema (gist) and progressively ``drills down'' to the Episodic Stream and Sensory Buffer only when decision uncertainty is high, retrieving fine-grained perceptual traces (verbatim) as needed.

To assess this paradigm, we conduct extensive evaluations on 4 challenging benchmarks, covering both offline long-video understanding and online streaming settings. Empirical results demonstrate that \modelname{} not only achieves a new state-of-the-art among open-source MLLMs and agentic systems by notable margins, but also exhibits competitive reasoning capabilities against proprietary models. Qualitative analyses and memory topology visualizations further reveal that \modelname{} can successfully decouple semantic ``gist'' from visual ``verbatim'' details, allowing the agent to perform precise detail verification without succumbing to the cognitive overload typical of vision-centric methods. These findings may inspire the development of robust and generalizable cognitive infrastructure for long-horizon autonomous agents.

Our contributions are summarized as follows:
\vspace{-0.5em}

\begin{itemize}[leftmargin=*]
    \item We propose \modelname{}, a pyramidal multimodal memory architecture grounded in Fuzzy-Trace Theory that bridges the gap between fine-grained perception and high-level cognition.
    \item We introduce SIB-GRPO, grounded in the Information Bottleneck principle, to optimize bottom-up memory construction by distilling knowledge from redundancy.
    \item We design an entropy-driven top-down memory retrieval strategy that adaptively ``drills down'' from schemas to details under high uncertainty, ensuring efficient and precise verification.
    \item Extensive experiments on four benchmarks demonstrate that \modelname{} achieves state-of-the-art performance and robust generalization across both offline and streaming scenarios.
\end{itemize}

%% file: sections/relatedwork.tex
\section{Related Work}
\label{sec:relatedwork}

\subsection{Long Video Understanding}

While multimodal large language models (MLLMs) have substantially extended vision-language capabilities from images to videos \cite{llava, blip, qwen3vl, LLaVA-Video}, long-video understanding remains constrained by limited context windows. Existing solutions generally fall into two paradigms.
\emph{Vision-centric} methods enhance context via dense sampling or token compression \cite{longva, Llama-vid, Ma-lmm}, with some incorporating auxiliary textual evidence \cite{VideoRAG}. Despite improved visual coverage, these approaches often suffer from high computational redundancy and inefficiency.
Conversely, \emph{Text-centric} approaches convert videos into captions or structured textual memories for efficiency \cite{ChatVideo, VideoTree, VideoMiner, Vgent, videoagent}. However, such conversion inevitably discards fine-grained visual cues and weakens perceptual grounding, thereby degrading complex reasoning over subtle details.

To bridge this gap, we propose a hierarchical pyramidal multimodal memory that unifies high-level textual memory for coarse localization and low-level visual memory for fine-grained retrieval, achieving a better balance between efficiency and visual fidelity~\cite{gao1}.

\subsection{Memory for Agents}

Memory mechanisms have been widely studied in agents built upon large language models (LLMs). Existing approaches span cache-like hierarchical designs \cite{MemGPT}, forgetting-curve-inspired memory management \cite{Memorybank}, associative memory linking \cite{A-mem}, and reinforcement learning-based memory control \cite{Memory-R1}. However, these systems, including LicoMemory \cite{licomemory}, remain largely \emph{text-centric}, limiting their ability to align information across modalities and to preserve the rich structural properties of real-world experiences.

In contrast, memory systems for \emph{multimodal agents} remain relatively underexplored. Prior work such as M3-Agent \cite{M3-Agent} typically relies on predefined memory structures and fixed operational workflows, which may constrain generalization in open-ended, long-horizon environments. These limitations highlight the need for a flexible and generalizable memory framework that can support long-term multimodal interactions.

%% file: sections/method.tex
\section{Method}
\label{sec:method}

\begin{figure*}[t]
    \centering
    \includegraphics[width=\linewidth]{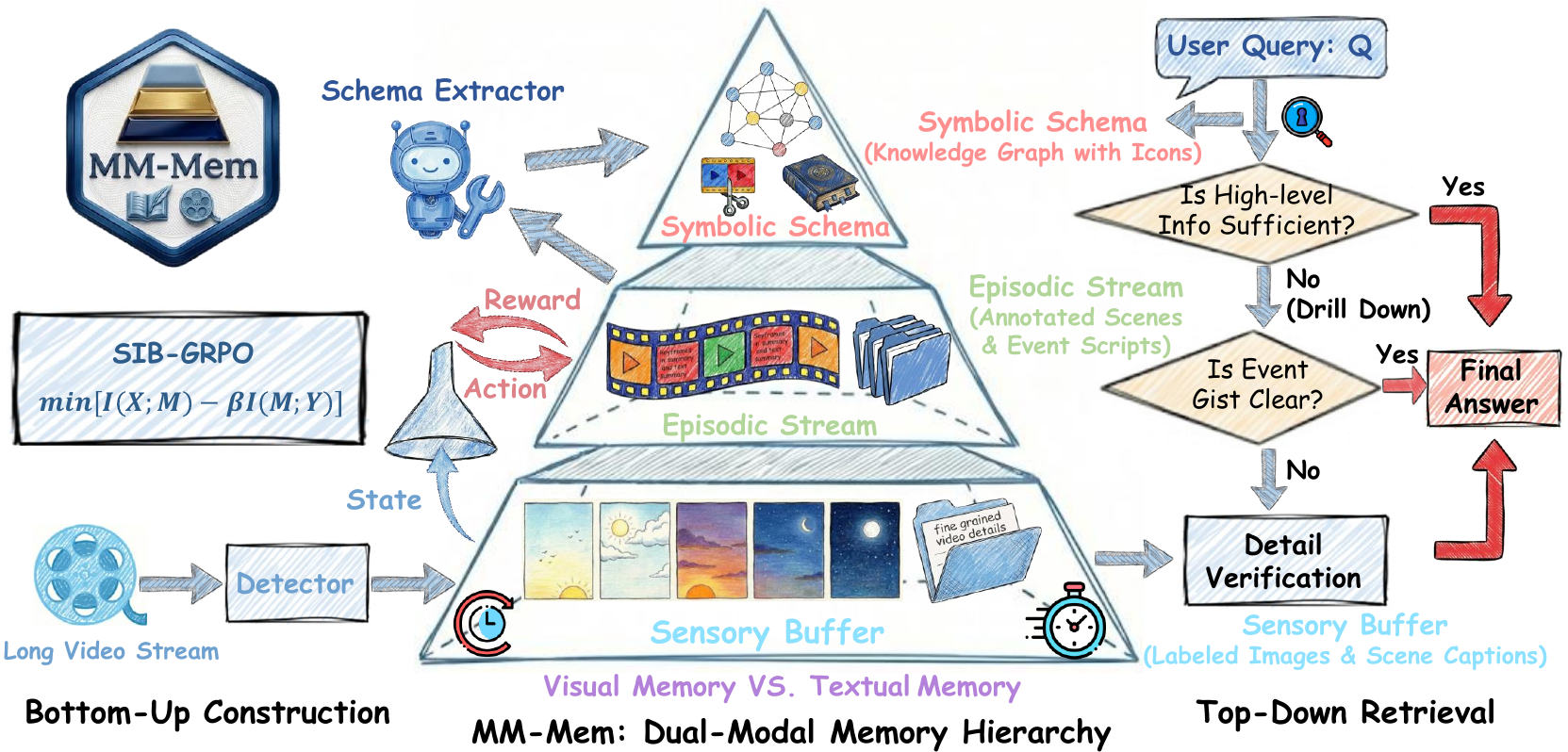}
    \caption{Overview of \modelname{}: \modelname{} unifies visual and textual memory through (left) a bottom-up memory construction process, which transforms raw sensory frames into abstract symbolic schemas, and (right) a top-down retrieval process that supports query-adaptive reasoning.}
    \label{fig:pipeline}
\end{figure*}

We propose \modelname{}, a multimodal memory architecture that helps agents perceive and understand the world. \modelname{} uses a bottom-up, offline-built hierarchical memory pyramid, spanning from a perception level \textit{Sensory Buffer} that retains fine-grained visual evidence to a semantics level \textit{Symbolic Schema} that stores high level textual abstractions. For long-horizon interaction, we introduce SIB-GRPO for dynamic memory management, which removes redundant memories while preserving task relevant semantics. We further design a top-down hierarchical retrieval mechanism guided by predictive entropy, which adaptively selects the retrieval depth to balance evidence coverage and resource constraints.

\subsection{Multimodal Pyramid Memory Structure}

Rather than physically decoupling modalities into isolated tracks, our three-layer hierarchy maintains integrated multimodal representations across all levels. While we conceptually map visual data to verbatim traces and textual data to gist traces inspired by FTT, both modalities coexist throughout the pyramid. The memory construction evolves through a representational shift: it begins as a vision-dominant multimodal memory at the bottom for fine-grained perception, and progressively distills into a text-dominant representation at the top for high-level cognitive abstraction.

\paragraph{Sensory Buffer}
Given a long video stream $\mathcal{V}$, we apply content-adaptive temporal segmentation to obtain clips $\mathcal{C}=\{c_t\}$ (e.g., PySceneDetect). For each clip $c_t$, we identify salient temporal indices $\mathcal{S}_t$ based on inter-frame variation and construct short key sub-clips centered at these indices (details in Appendix~\ref{app:sensory-keyframe}). Sensory memory is instantiated as
\begin{equation}
\mathcal{M}_{\text{sens}}
=\{(\mathbf{v}_{t,i},\mathbf{l}_{t,i},\tau_{t,i}) \mid i\in\mathcal{S}_t,\; c_t\in\mathcal{C}\},
\end{equation}
where $\mathbf{v}_{t,i}$ is the visual representation, $\mathbf{l}_{t,i}$ is the associated text trace (e.g., subtitles or clip captioning), and $\tau_{t,i}$ is the temporal location (e.g., center-frame timestamp). Crucially, at this vision-dominant foundational layer, the text trace $\mathbf{l}_{t,i}$ is strongly bound to its visual counterpart and functions as an auxiliary component. Rather than acting as an independent or high-level semantic representation, the text serves purely as a descriptive label for visual entities. It provides a semantic anchor to help index and isolate the dense, highly redundant verbatim visual details.

\paragraph{Episodic Stream}
Following selective encoding and temporal contiguity \cite{memory1}, we construct an \emph{Episodic Stream} by consolidating sensory entries from $\mathcal{M}_{\text{sens}}$.
Each sensory item is $m_{t,i}=(\mathbf{v}_{t,i},\mathbf{l}_{t,i},\tau_{t,i})\in\mathcal{M}_{\text{sens}}$.
We maintain an ordered episodic sequence $\mathcal{M}_{\text{epi}}$, with the latest retained node $e^{\star}=(\mathbf{e}^{\star},\mathbf{l}^{\star},\tau^{\star})$.

For each $m_{t,i}$, a decision operator $\psi(\cdot)$ updates the stream:
\begin{equation}
a_{t,i}=\psi(m_{t,i},e^{\star}),\qquad a_{t,i}\in\mathcal{O},
\end{equation}
where $\mathcal{O}=\{\textsc{ADD\_NEW},\textsc{MERGE},\textsc{DISCARD}\}$.
\textsc{ADD\_NEW} appends a node initialized from $m_{t,i}$; \textsc{MERGE} integrates $m_{t,i}$ into $e^{\star}$; and \textsc{DISCARD} removes redundant or low-novelty items. A single chronological pass over $\mathcal{M}_{\text{sens}}$ yields a compact episodic stream.

To obtain event-level abstractions, we cluster retained visual representations (e.g., via $K$-means) and select representative prototypes as summaries. The resulting episodic memory is
\begin{equation}
\mathcal{M}_{\text{epi}}=\{(\mathbf{e}_k,\mathbf{l}_k,\tau_k)\}_{k=1}^{|\mathcal{M}_{\text{epi}}|},
\end{equation}
where $\mathbf{e}_k$ is the representation of the $k$-th episodic unit, $\mathbf{l}_k$ aggregates its associated textual traces, and $\tau_k$ records its temporal span.

\paragraph{Symbolic Schema}
To support cross-episode reasoning \cite{memory1}, we build a \emph{Symbolic Schema} as a knowledge graph $\mathcal{G}=(\mathcal{N},\mathcal{E})$ over episodic memory $\mathcal{M}_{\text{epi}}$. An MLLM extracts entities and glosses from each episodic unit $e_k \in \mathcal{M}_{\text{epi}}$, unifying them into a global prototype set $\mathcal{U}$ with aggregated glosses $t_u$. 

The graph nodes $\mathcal{N} = \{v_k\} \cup \mathcal{U}$ comprise episodic units and prototypes. The edges $\mathcal{E}$ contain optional semantic relations $u_p\xrightarrow{r}u_q$ and, crucially, grounding edges $(v_k,u)$. These grounding edges serve as explicit multimodal pointers: rather than collapsing into a unimodal textual graph, they tightly anchor text-dominant concepts (semantic gist) back to specific episodic units (retaining verbatim visual evidence). 

Thus, symbolic memory is instantiated as a text-driven multimodal index:
\begin{equation}
\mathcal{M}_{\text{sym}}=\{(u,t_u,\mathcal{P}_u)\}_{u\in\mathcal{U}},
\end{equation}
where $\mathcal{P}_u = \{v_k \mid (v_k, u) \in \mathcal{E}\}$ denotes the visual pointers for concept $u$. This design enables efficient high-level cognitive reasoning while preserving dynamic drill-down to concrete visual details.

\subsection{Bottom-Up Memory Construction}

A bottom-up pipeline transforms raw videos into a three-level memory hierarchy: \textit{Sensory Buffer}, \textit{Episodic Stream}, and \textit{Symbolic Schema}. Fine-grained perceptual signals are retained in the \textit{Sensory Buffer}; segments are organized into compact event traces in the \textit{Episodic Stream}; and structured knowledge is consolidated in the \textit{Symbolic Schema} over longer time scales.

During Sensory-to-Episodic construction, redundancy must be compressed while preserving task-relevant semantics and controlling memory growth. We introduce SIB-GRPO to fine-tune the memory manager with reinforcement learning, enabling adaptive generation of information-dense episodic traces. Crucially, driven by SIB-GRPO, this bottom-up construction is not merely a process of data reduction; it orchestrates a smooth transition in modality dominance. The system progressively distills high-information-density textual gist from the heavily redundant visual verbatim traces, naturally aligning with the principles of Fuzzy-Trace Theory. We next describe the Sensory-to-Episodic pipeline and optimization.

\subsubsection{Sensory-to-Episodic Memory}
Given a sensory buffer $\mathcal{M}_{\text{sens}}=\{m_{t,i}=(\mathbf{v}_{t,i},\mathbf{l}_{t,i},\tau_{t,i})\}$, we construct a compact episodic stream $\mathcal{M}_{\text{epi}}$ that retains task-relevant semantics for downstream reasoning while discarding redundant, low-novelty details.
Let $X$ denote the sensory memory content (a temporally local window from $\mathcal{M}_{\text{sens}}$ together with the latest episodic node $e^{\star}$), and let $M$ denote the episodic representation produced by a stochastic encoder (memory manager) $p_{\theta}(m\mid x)$. 
We cast this \emph{Sensory-to-Episodic} conversion as stochastic compression, where a memory manager serves as an encoder mapping sensory $X$ to episodic memory $M$.

\paragraph{Remark: Action-output correspondence}
Under fixed update rules, $\mathcal{O}$ uniquely determines $M$ (see Appendix~\ref{abb:remark}).

\paragraph{Semantic IB Formulation}
Let $X$ denote the sensory memory content (a temporally local window from $\mathcal{M}_{\text{sens}}$ together with the latest episodic node $e^{\star}$), and let $M$ denote the episodic representation produced by a stochastic encoder (memory manager) $p_{\theta}(m\mid x)$.
We adopt an Information Bottleneck (IB) objective \cite{Bottleneck}:

\vspace{-0.5em}

\begin{equation}
\min_{p_{\theta}(m\mid x)}\ \big[I(X;M)-\beta\,I(M;Y)\big],
\end{equation}
where $Y$ is the supervision label (ground-truth VQA answer) and $\beta$ controls the compression--prediction trade-off. Here $I(\cdot;\cdot)$ denotes the mutual information between two random variables.

To obtain a tractable training objective, we introduce a variational decoder $q_{\phi}(y\mid m)$ to approximate $p_{\theta}(y\mid m)$ and a variational prior $r(m)$ to approximate the marginal $p_{\theta}(m)$. We define
\begin{equation}
\mathcal{L}_{\text{p}}(\theta,\phi)
\triangleq
\mathbb{E}_{p(x,y)\,p_{\theta}(m\mid x)}
\big[\log q_{\phi}(y\mid m)\big],
\end{equation}
\begin{equation}
\mathcal{L}_{\text{c}}(\theta)
\triangleq
\mathbb{E}_{p(x)}
\Big[D_{\mathrm{KL}}\big(p_{\theta}(m\mid x)\,\|\,r(m)\big)\Big].
\end{equation}
As shown in Appendix~\ref{app:proof}, $\mathcal{L}_{\text{p}}$ lower-bounds $I(M;Y)$ up to an additive constant $H(Y)$, and $\mathcal{L}_{\text{c}}$ upper-bounds $I(X;M)$. Therefore, dropping the constant $H(Y)$, we optimize the following variational IB objective:
\begin{equation}
\max_{\theta,\phi}\ \beta\,\mathcal{L}_{\text{p}}(\theta,\phi)\ -\ \mathcal{L}_{\text{c}}(\theta).
\end{equation}

\vspace{-0.3em}

\paragraph{A Quality--Quantity Prior}
To encode an explicit \emph{quality--quantity} trade-off in episodic memory, we adopt the prior
\begin{equation}
r(m)\ \propto\ \underbrace{p_{\text{ref}}(m)}_{\text{Quality}}\ \cdot\ \underbrace{e^{-\lambda|m|}}_{\text{Quantity}},
\end{equation}
where $\lambda$ is a hyperparameter, $|m|$ is the token length of textual trace at an \emph{Episodic Stream} node, and $p_{\text{ref}}$ is a \emph{teacher} distribution promoting fluent, general-purpose memory expressions while anchoring the policy to a trusted prior. This factorization yields an IB regularizer consisting of a length penalty and a KL term to the teacher, resembling RLHF-style trust-region objectives.

\paragraph{SIB-GRPO: Dynamic Management}
While the IB objective provides a principled semantic compression criterion, the episodic trace is generated discretely (LLM-style) and must be optimized with sequence-level feedback. We therefore train the Memory Manager as a policy $\pi_{\theta}(m\mid s)$ using reinforcement learning, where $s=(x,M_{\text{old}})$ and the action is a textual trace $m$ appended to $\mathcal{M}_{\text{epi}}$.

For each sampled $m$, we compute designed scalar reward
\begin{equation}
\begin{aligned}
r(s,m)
&\triangleq
\underbrace{R_{\text{vqa}}(s,m)}_{\text{Task Reward}}
\;-\;
\underbrace{\beta_1 \cdot \text{Length}(m)}_{\text{from } e^{-\lambda |m|}} \\
&\quad\quad\quad
-\;
\underbrace{
\beta_2 \cdot
\log \frac{\pi_{\theta_{\text{old}}}(m\mid s)}{\pi_{\text{ref}}(m\mid s)}
}_{\text{from } p_{\text{ref}}(m)} .
\end{aligned}
\end{equation}

Here $\pi_{\text{ref}}$ (equivalently $p_{\text{ref}}$) is a fixed reference policy/distribution. In practice, the log-ratio is evaluated on sampled $m$ from the behavior policy $\pi_{\theta_{\text{old}}}$ and serves as a KL-style regularizer that anchors $\pi_{\theta}$ to $\pi_{\text{ref}}$.

Given a state $s=(x,M_{\text{old}})$, we sample a group of $G$ candidate episodic traces $\{m_i\}_{i=1}^{G}\sim \pi_{\theta_{\text{old}}}(\cdot\mid s)$, compute their scalar rewards $\{r_i\}_{i=1}^{G}$, and construct a standardized group-relative advantage $A_i$ (computed by normalizing each $r_i$ within the group). We then optimize a PPO-style clipped surrogate using the importance ratio $\rho_i(\theta)=\pi_{\theta}(m_i\mid s)/\pi_{\theta_{\text{old}}}(m_i\mid s)$, where $\epsilon$ controls the clipping range. The resulting SIB-GRPO objective is
\begin{equation}
\begin{aligned}
J_{\text{SIB-GRPO}}(\theta)
&=
\mathbb{E}_{s,\{m_i\}\sim \pi_{\theta_{\text{old}}}}\Bigg[
\frac{1}{G}\sum_{i=1}^{G}\min 
\\
&\hspace{-5.3em}
\Bigg(\rho_i(\theta)A_i,\;
\mathrm{clip}\!\big(\rho_i(\theta),\,1-\epsilon,\,1+\epsilon\big)\,A_i\Bigg)\Bigg],
\end{aligned}
\end{equation}
In practice, we minimize $-J_{\text{SIB-GRPO}}(\theta)$ as training loss.

\subsection{Entropy-Driven Top-Down Retrieval}

A top-down, coarse-to-fine retrieval strategy is adopted, querying memory from high-level semantic abstractions to progressively finer perceptual evidence. Retrieval begins at the \textit{Symbolic Schema} level via text to rapidly instantiate the semantic gist. If uncertainty persists, the query descends to lower layers and may terminate at the \textit{Episodic} layer once sufficient evidence is obtained; only when ambiguity remains is it routed to the \textit{Sensory Buffer}, retrieving CLIP-encoded visual keyframes to resolve decisions with local visual details.

This design follows Reverse Hierarchy Theory \cite{RHT}, which posits that perception is initiated with high-level \emph{vision at a glance} and refined by low-level \emph{vision with scrutiny} when fine-grained discrimination is required.

Formally, given a question $\mathcal{Q}$ and a candidate answer set $\mathcal{A}=\{a_1,\ldots,a_N\}$, each retrieval step $s$ returns evidence $R_s$.
Given accumulated evidence $R_{\le s}$, a posterior distribution $p(a_i \mid \mathcal{Q}, R_{\le s})$ is maintained over candidates.
For notational convenience, let
\begin{equation}
p_i^{(s)} \triangleq p(a_i \mid \mathcal{Q}, R_{\le s}).
\end{equation}
The entropy of this answer distribution is used as an adaptive stopping criterion:
\begin{equation}
H_s(\mathcal{Q}) = -\sum_{i=1}^{N} p_i^{(s)} \log p_i^{(s)}.
\end{equation}
Retrieval is terminated when $H_s(\mathcal{Q}) \le \gamma$, or when the entropy reduction $\Delta H_s = H_{s-1}(\mathcal{Q}) - H_s(\mathcal{Q})$ falls below a small $\epsilon$ for several consecutive steps, and the most probable answer is returned:
$\arg\max_{a_i \in \mathcal{A}} p_i^{(s)}$.
Intuitively, rapid semantic narrowing of $\mathcal{A}$ is enabled by high-level text retrieval, whereas low-level keyframe retrieval is invoked only under high entropy, yielding a compute--accuracy trade-off that adapts to question difficulty.

\begin{table*}[t]
\centering
\resizebox{0.9\textwidth}{!}{
\begin{tabular}{lllllllllc}
\noalign{\hrule height 1.2pt}

\multicolumn{1}{c}{\multirow{3.8}{*}{\textbf{Method}}} 
& \multicolumn{8}{c}{\textbf{Video-MME}} 
& \multicolumn{1}{c}{\textbf{MLVU}} \\

\cmidrule(lr){2-9}
\cmidrule(lr){10-10}

& \multicolumn{2}{c}{\textbf{Short}}              
& \multicolumn{2}{c}{\textbf{Medium}}             
& \multicolumn{2}{c}{\textbf{Long}}               
& \multicolumn{2}{c}{\textbf{Overall}}            
& \multicolumn{1}{c}{\multirow{2}{*}{\textbf{M-Avg}}} \\

\cmidrule(lr){2-3}
\cmidrule(lr){4-5}
\cmidrule(lr){6-7}
\cmidrule(lr){8-9}

\multicolumn{1}{c}{}                                
& \multicolumn{1}{c}{w/o} & \multicolumn{1}{c}{w/} 
& \multicolumn{1}{c}{w/o} & \multicolumn{1}{c}{w/} 
& \multicolumn{1}{c}{w/o} & \multicolumn{1}{c}{w/} 
& \multicolumn{1}{c}{w/o} & \multicolumn{1}{c}{w/} 
& \\ \hline \hline
\rowcolor{blue!4}
\multicolumn{10}{c}{\textit{Proprietary Models}}    \\ 
Gemini 1.5 Pro \cite{Gemini}                                     & 81.7                    & 84.5                  & 74.3                    & 81.0                    & 67.4                    & 77.4                  & 75.0                      & 81.3                  & \multicolumn{1}{c}{-} \\
GPT-4o     \cite{GPT-4O}                                         & 80.0                      & 82.8                  & 70.3                    & 76.6                  & 65.3                    & 72.1                  & 71.9                    & 77.2                  & \multicolumn{1}{c}{64.6} \\
Gemini 1.5 Flash   \cite{Gemini}                                  & 78.8                    & 79.8                  & 68.8                    & 74.7                  & 61.1                    & 68.8                  & 70.3                    & 75.0                    & \multicolumn{1}{c}{-} \\
GPT-4V    \cite{GPT-4}                                          & 70.5                    & 73.2                  & 55.8                    & 59.7                  & 53.5                    & 56.9                  & 59.9                    & 63.3                  & \multicolumn{1}{c}{49.2} \\ \hline \hline
\rowcolor{blue!4}
\multicolumn{10}{c}{\textit{Open-Sourced MLLMs}}   \\ 
Qwen2-VL-72B      \cite{qwen2-vl}                                  & 80.1                    & 82.2                  & 71.3                    & 76.8                  & 62.2                    & 74.3                  & 71.2                    & 77.8                  & \multicolumn{1}{c}{-} \\
LLaVA-Video-72B  \cite{LLaVA-Video}                                   & 81.4                    & 82.8                  & 68.9                    & 75.6                  & 61.5                    & 72.5                  & 70.6                    & 76.9                  & \multicolumn{1}{c}{73.1} \\
Qwen3-VL-8B      \cite{qwen3vl}                                   & 76.4                   &  79.2                     & 62.3                   & 72.6                      & 55.9                   &      70.8                 & 64.9                   &   74.2                    & \multicolumn{1}{c}{65.9} \\
VideoLLaMA 3-7B  \cite{Videollama3}                                   & 80.1                    & 80.2                  & 63.7                    & 69.6                  & 54.9                    & 61.0                    & 66.2                    & 70.3                  & \multicolumn{1}{c}{73.0} \\

VideoLLaMA 2-72B \cite{Videollama2}                                   & 69.8                    & 72.0                    & 59.9                    & 63.0                    & 57.6                    & 59.0                    & 62.4                    & 64.7                  & \multicolumn{1}{c}{-} \\
VITA 1.5-7B  \cite{Vita-1.5}                                       & 67.0                      & 69.9                  & 54.2                    & 55.7                  & 47.1                    & 50.4                  & 56.1                    & 58.7                  & \multicolumn{1}{c}{60.4} \\
Long-LLaVA-7B  \cite{longllava}                                     & 61.9                    & 66.2                  & 51.4                    & 54.7                  & 45.4                    & 50.3                  & 52.9                    & 57.1                  & \multicolumn{1}{c}{-} \\
LongVA-7B  \cite{longva}                                         & 61.1                    & 61.6                  & 50.4                    & 53.6                  & 46.2                    & 47.6                  & 52.6                    & 54.3                  & \multicolumn{1}{c}{-} \\
Video-LLaVA-7B   \cite{Video-llava}                                   & 45.3                    & 46.1                  & 38.0                      & 40.7                  & 36.2                    & 38.1                  & 39.9                    & 41.6                  & \multicolumn{1}{c}{-} \\ \hline \hline
\rowcolor{blue!4}
\multicolumn{10}{c}{\textit{Agent-based Systems}}       \\ 
Vgent    \cite{Vgent}                                           & \multicolumn{1}{c}{-}   & \multicolumn{1}{c}{-} & \multicolumn{1}{c}{-}   & \multicolumn{1}{c}{-} & \multicolumn{1}{c}{-}   & \multicolumn{1}{c}{-} & 68.9                    & 74.3                  & \multicolumn{1}{c}{72.1} \\ 
VideoRAG     \cite{VideoRAG}                                       & \multicolumn{1}{c}{-}   & 67.1                  & \multicolumn{1}{c}{-}   & 60.4                  & \multicolumn{1}{c}{-}   & 60.1                  & 60.5                    & 62.6                  & \multicolumn{1}{c}{72.4} \\
VideoMiner   \cite{VideoMiner}                                       & 65.6                    & \multicolumn{1}{c}{-} & 57.5                    & \multicolumn{1}{c}{-} & 52.2                    & \multicolumn{1}{c}{-} & 58.4                    & \multicolumn{1}{c}{-} & \multicolumn{1}{c}{65.1} \\
VideoTree      \cite{VideoTree}                                     & 55.5                    & \multicolumn{1}{c}{-} & 49.2                    & \multicolumn{1}{c}{-} & 39.3                    & \multicolumn{1}{c}{-} & 48.0                      & \multicolumn{1}{c}{-} & \multicolumn{1}{c}{60.4} \\

\rowcolor{purple!7}
\textbf{\modelname{} (Ours)}  & $\bm{81.5}$ & $\bm{82.8}$ & $\bm{69.6}$ & $\bm{75.8}$ &
$\bm{66.1}$ & $\bm{75.7}$ & $\bm{72.4}$ & $\bm{78.1}$ & $\bm{77.2}$ \\ \hline \hline
\end{tabular}
}
\caption{Comparison on two long-video understanding benchmarks: \textbf{\textsc{Video-MME}} and \textbf{\textsc{MLVU}}. For \textsc{Video-MME}, we report results under both settings (w/ = with subtitles; w/o = without subtitles). For \textsc{MLVU}, we report M-Avg. Most baseline results are taken from the official leaderboards (as of 2026-01-01) or the respective papers. Results not reported in the original sources are marked with ``--``. For agent-based systems, we report the best-performing configuration with a comparable parameter scale.}
\label{tab:long-video}
\end{table*}

%% file: sections/experiments.tex
\section{Experiment}
\label{sec:experiment}

\begin{figure*}[t]
    \centering
    \begin{subfigure}[t]{0.30\textwidth}
        \centering
        \includegraphics[width=\linewidth]{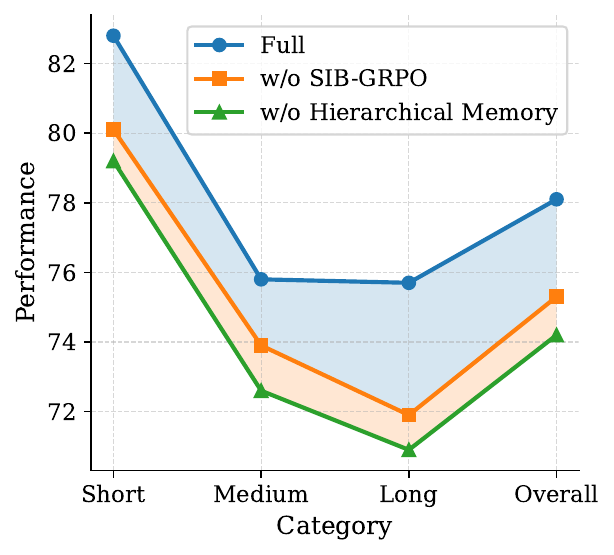}
        \caption{Ablation results.}
        \label{fig:ablation}
    \end{subfigure}
    \hfill
    \begin{subfigure}[t]{0.33\textwidth}
        \centering
        \includegraphics[width=\linewidth]{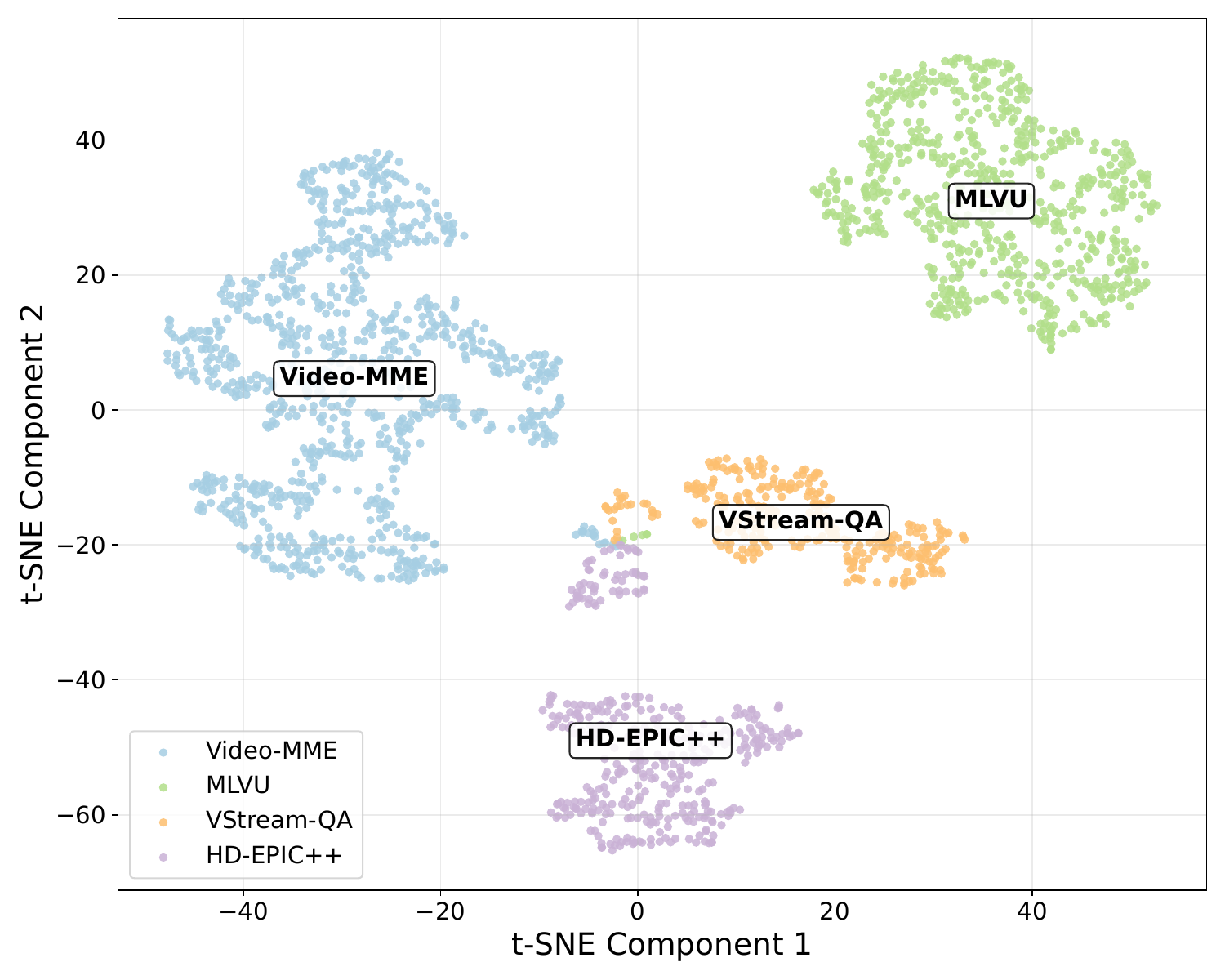}
        \caption{Sensory Buffer Visual Memory.}
        \label{fig:mem1}
    \end{subfigure}
    \hfill
    \begin{subfigure}[t]{0.33\textwidth}
        \centering
        \includegraphics[width=\linewidth]{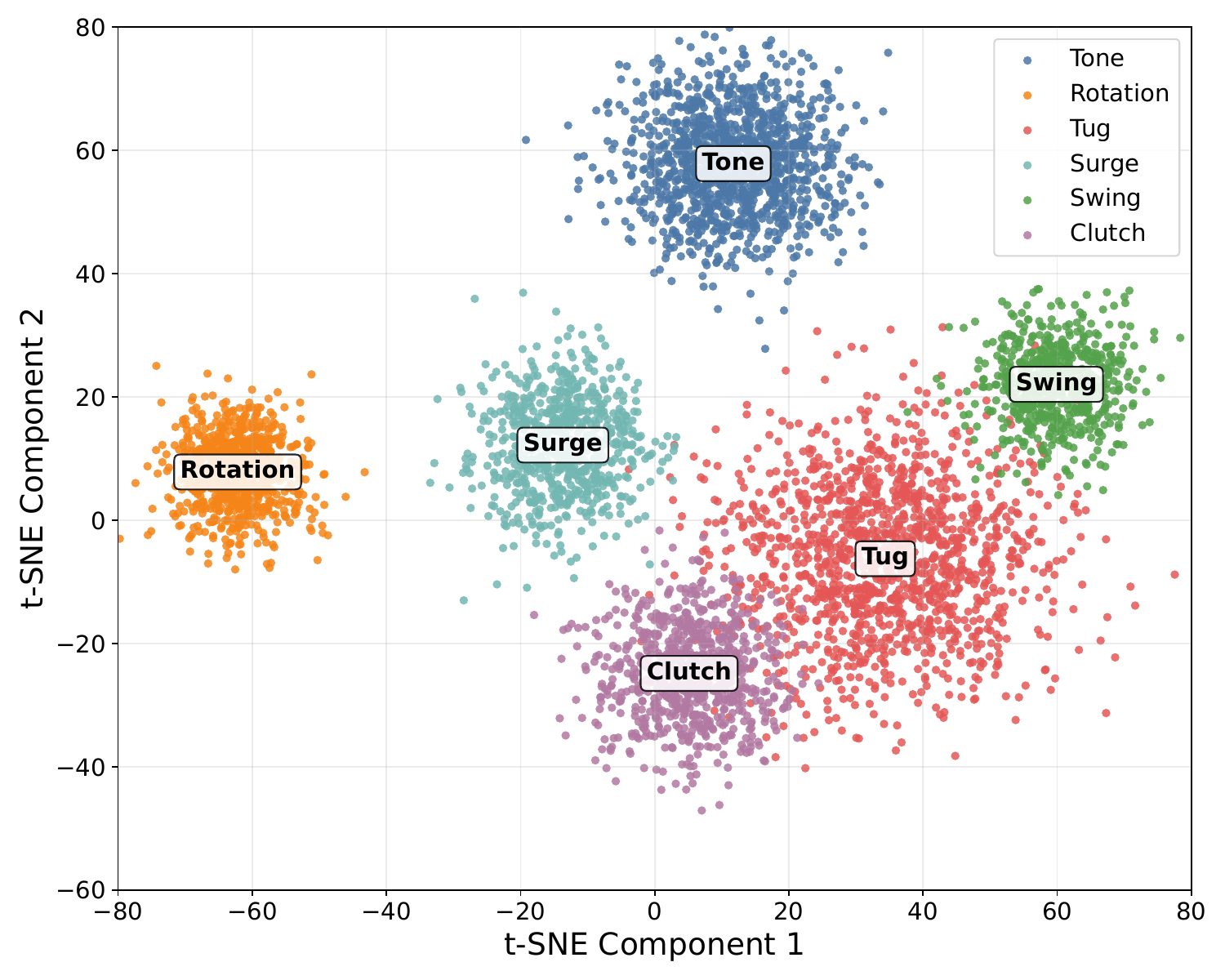}
        \caption{Episodic Stream Visual Memory}
        \label{fig:mem2}
    \end{subfigure}
    \caption{Visualization of ablation results and memory representations.}
    \label{fig:three_in_one}
\end{figure*}

\subsection{Experimental Setup}

\vspace{-0.5em}
\paragraph{Benchmarks}
Four benchmarks are comprehensively evaluated.
\textbf{(i) Standard long-video datasets.}
\textsc{Video-MME} \cite{Video-MME} covers three length regimes, namely short ($<2$ min), medium (4--15 min), and long (30--60 min), with 300 videos each (900 total) and 2,700 questions; we report both \emph{with-subtitle} and \emph{without-subtitle} settings.
\textsc{MLVU} \cite{MLVU} \textbf{dev set} includes nine tasks with videos from 3 min to 2 h (avg. $\sim$12 min).
\textbf{(ii) Standard online streaming dataset.}
\textsc{VStream-QA} \cite{VStream-QA} contains \textsc{VStream-QA-Ego} and \textsc{VStream-QA-Movie} for egocentric and third-person narrative understanding.
\textbf{(iii) A derived egocentric long-video dataset.}
We built \textsc{HD-EPIC++} from \textsc{HD-EPIC} \cite{HD-EPIC} by re-splitting train/test, comprising 156 videos; details see Appendix \ref{app:hdepicpp}.

\vspace{-0.5em}
\paragraph{Evaluation protocol}
For comparison, we follow baseline settings.
For \textsc{Video-MME}, \textsc{MLVU}, and \textsc{HD-EPIC++}, accuracy is used as the evaluation metric.
Since \textsc{VStream-QA} consists of open-ended questions, \textsc{gpt-4o-mini} \cite{GPT-4O} is leveraged as an automatic judge. We report both accuracy and the averaged score on \textsc{VStream-QA}.
Implementation details and evaluation scripts are provided in Appendix \ref{abb:eval}.

\vspace{-0.5em}
\paragraph{Implementation details}
Experiments are conducted on NVIDIA A100 80GB GPUs.
\modelname{} uses Qwen3-VL-8B \cite{qwen3vl} as the base model.
For text retrieval, we use \textsc{bge-large-en-v1.5} \cite{BGE-1} and \textsc{bge-reranker-v2-m3} \cite{BGE-2}.
For visual retrieval, we use \emph{clip-level} retrieval by jointly scoring keyframes per clip; CLIP-style embeddings come from the base model's vision encoder.
Models are served with vLLM, and fine-tuning is performed under SWIFT with \textsc{SIB-GRPO}.
We set $\beta_1=0.1$, $\beta_2=0.3$, and temperature to $0.0$.
Hyperparameters are provided in Appendix \ref{abb:hyper}.

\subsection{Comparison with State-of-the-arts}

\vspace{-0.5em}
\paragraph{Long Video Understanding}
Baselines follow each benchmark leaderboard and common protocols in prior work, covering (a) proprietary multimodal models, (b) open-source MLLMs, and (c) agent-based systems for long-horizon video understanding. As shown in Table~\ref{tab:long-video}, \modelname{} consistently outperforms prior agent systems. Compared with the strongest agent baseline \textsc{Vgent}, \modelname{} yields a $\bm{5.1\%}$ \emph{relative} gain on \textsc{Video-MME} (both w/o- and w/-subtitle) and a $\bm{7.1\%}$ gain on \textsc{MLVU} in \textbf{M-Avg}. Despite using \textsc{Qwen3-VL-8B} as the backbone, \modelname{} surpasses all compared open-source MLLMs (e.g., \textsc{Qwen2-VL-72B}) and is competitive with strong proprietary models such as \textsc{Gemini 1.5 Pro}.

\begin{table}[t]
\centering
\resizebox{0.43\textwidth}{!}{
\begin{tabular}{l cc}
\toprule
\multicolumn{1}{c}{\multirow{2}{*}{\textbf{Method}}} & \multicolumn{2}{c}{\textbf{VStream-QA-Ego}} \\
\cmidrule(lr){2-3}
                                & \textbf{Accuracy} & \textbf{Score} \\   \hline
\midrule
Video-ChatGPT \cite{Video-ChatGPT}                   & 51.7 & 3.7 \\
MovieChat \cite{MovieChat}                      & 52.2 & 3.4 \\
Chat-UniVi \cite{Chat-UniVi}                      & 50.9 & 3.8 \\
LLaMA-VID  \cite{Llama-vid}                     & 54.8 & 3.9 \\
Flash-VStream    \cite{VStream-QA}               & 59.0 & 3.9 \\
\rowcolor{purple!7}
\textbf{\modelname{} (Ours) }                         & $\bm{62.5}$ & $\bm{4.1}$ \\ \hline
\bottomrule
\end{tabular}
}
\caption{Test performance on \textbf{\textsc{VStream-QA-Ego}}.}
\label{tab:VStream-QA}
\end{table}

\paragraph{Online Streaming Video Understanding}
Unlike most prior work that evaluates only on long-video benchmarks, long-video understanding primarily targets an offline setting, where the model is provided with the user query and a single video clip of finite length at the same time. To better approximate real-world online video-stream scenarios, we additionally evaluate on the streaming benchmark \textsc{VStream-QA-Ego}. As shown in Table~\ref{tab:VStream-QA}, \modelname{} remains effective for long-horizon streaming inputs, improving over the previous best method \textsc{Flash-VStream} by $\bm{5.9\%}$ and $\bm{5.2\%}$ in terms of \textbf{Accuracy} and \textbf{Score}, respectively.

\begin{table}[t]
\centering
\resizebox{0.4\textwidth}{!}{
\begin{tabular}{l l}
\hline
\multicolumn{1}{c}{\multirow{2}{*}{\textbf{Method}}}
 & \multicolumn{1}{c}{\textbf{HD-EPIC++}} \\ \cline{2-2}
                                   & \multicolumn{1}{c}{\textbf{Accuracy}}  \\ \hline \hline
Qwen3-VL-8B \cite{qwen3vl}          & \multicolumn{1}{c}{25.88} \\
Qwen2.5-VL-7B  \cite{qwen3vl}      & \multicolumn{1}{c}{24.37} \\
LLaVA-Video-7B  \cite{qwen3vl}     & \multicolumn{1}{c}{25.37} \\
VideoLLaMA 3-7B \cite{Videollama3}  & \multicolumn{1}{c}{20.36} \\ 
Qwen3-VL-4B  \cite{qwen3vl}          & \multicolumn{1}{c}{24.91} \\
Qwen3-VL-2B  \cite{qwen3vl}         & \multicolumn{1}{c}{22.80}  \\
\rowcolor{purple!7}
\textbf{\modelname{} (Ours)}                              & \multicolumn{1}{c}{$\bm{30.28}$} \\ \hline \hline
\end{tabular}
}
\caption{Evaluation on the built \textbf{\textsc{HD-EPIC++}}.}
\label{tab:hd-epic++}
\end{table}

\paragraph{Egocentric Long Video Understanding}
Table~\ref{tab:hd-epic++} reports accuracy on \textsc{HD-EPIC++}. Our \modelname{} achieves \textbf{30.28\%}, outperforming all baselines. It exceeds the strongest competitor (Qwen3-VL-8B) by \textbf{+4.40} points (30.28 vs.\ 25.88), and also surpasses LLaVA-Video-7B and VideoLLaMA 3-7B by \textbf{+4.91} and \textbf{+9.92} points. This suggests \modelname{} better aggregates fine-grained egocentric cues over long temporal contexts.

\subsection{Ablation Studies}

\paragraph{Effectiveness of SIB-GRPO and Pyramid Memory}
Figure~\ref{fig:ablation} reports an ablation study over \textit{Short}, \textit{Medium}, \textit{Long}, and \textit{Overall} splits.
Our full model performs best across all categories, indicating positive contributions from each component.
Removing SIB-GRPO consistently degrades performance, with the largest drop on \textit{Long}, suggesting its importance for consolidation under long temporal dependencies.
Further removing the Pyramid (hierarchical) memory yields an additional decrease, again most pronounced on \textit{Long} and \textit{Overall}.
These results show that pyramid memory complements SIB-GRPO by organizing information at multiple temporal/semantic granularities, improving retention and retrieval for long-horizon reasoning.
Additional ablations and hyper-parameter analyses are provided in Appendix~\ref{abb:add_exp}.

\paragraph{Topology of the Cognitive Memory Space}
To qualitatively assess the structure of our hierarchical memory, we project memory embeddings to 2D with t-SNE. Figure~\ref{fig:mem1} (Middle) shows Sensory Buffer representations across benchmarks: the clear separation between egocentric (HD-EPIC++) and cinematic (Video-MME) domains indicates that the L1 layer preserves domain-specific visual details without collapse. Figure~\ref{fig:mem2} (Right) visualizes the Episodic Stream, where semantic clusters emerge naturally (e.g., `Rotation' vs.\ `Swing'), suggesting that RL-driven consolidation suppresses noise and promotes abstraction from sensory signals to higher-level reasoning.

\subsection{Efficiency and Deployment Analysis}
\label{sec:efficiency_videomme}

We analyze the efficiency of \modelname{} on \textbf{Video-MME} without subtitles, covering the Short, Medium, and Long splits. Since runtime is highly correlated with raw video duration, we normalize all timing metrics by the \emph{time required to process one minute of video}, ensuring fair and comparable evaluation. We report \textbf{Peak VRAM}, \textbf{Memory Construction Time}, and \textbf{Inference Latency}. Here, \textbf{N/A} denotes results that we are unable to reproduce due to closed-source implementations or insufficient implementation details.

\begin{table}[t]
\centering
\resizebox{0.5\textwidth}{!}{
\begin{tabular}{lccc}
\hline
\textbf{Method} & \textbf{VRAM(GB)} & \textbf{Con. Time(s)} & \textbf{Infer. Latency(s)} \\
\hline\hline
\rowcolor{blue!4}
\multicolumn{4}{c}{\textbf{Proprietary MLLMs}} \\
VideoAgent     & N/A  & N/A   & 67.25 \\
\hline\hline
\rowcolor{blue!4}
\multicolumn{4}{c}{\textbf{Open-Source MLLMs}} \\
Qwen3-VL-8B    & 22.8 & N/A   & 6.47  \\
Video-RAG      & 23.0 & N/A   & 25.93 \\
Vgent          & 18.7 & 20.18 & 7.38  \\
\rowcolor{purple!7}
\textbf{MM-Mem (Ours)} & \textbf{17.8} & \textbf{19.54} & \textbf{5.35} \\
\hline\hline
\end{tabular}
}

\caption{Efficiency comparison on Video-MME without subtitles. \textbf{Con. Time} denotes construction time, and \textbf{Infer. Latency} denotes inference latency. All timing metrics are normalized to the time required to process one minute of video.}
\label{tab:efficiency_videomme}
\end{table}

\begin{figure*}[t]
    \centering
    \includegraphics[width=\linewidth]{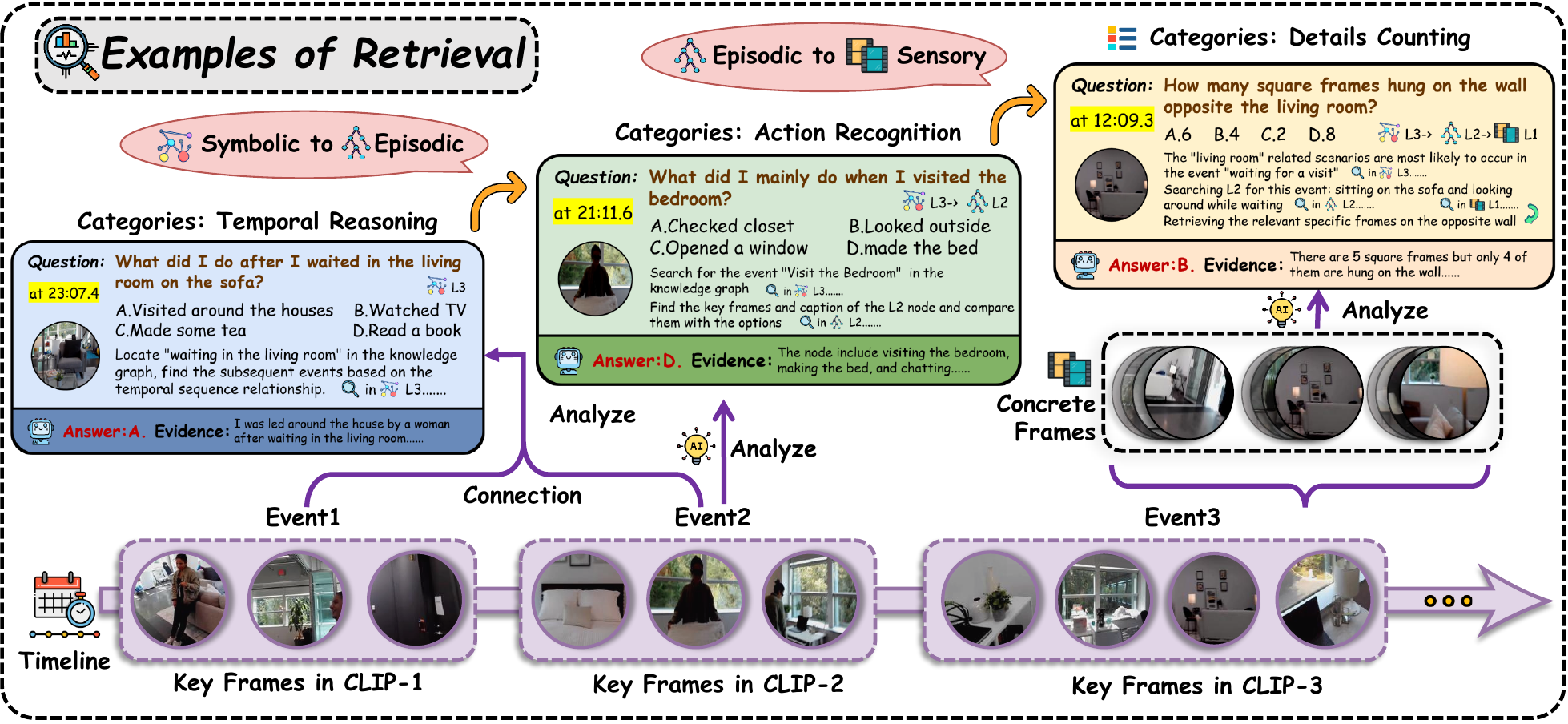}
    \caption{A qualitative example of \modelname{}'s coarse-to-fine retrieval across memory layers.}
    \label{fig:example}
\end{figure*}

\subsection{Qualitative Analysis}
\label{sec:qualitative}

As shown in Table~\ref{tab:efficiency_videomme}, \modelname{} achieves a favorable trade-off among construction cost, online latency, and GPU memory usage. Its offline memory construction can be amortized across multiple queries for the same video, which is especially beneficial in Video-MME-style settings. Meanwhile, \modelname{} supports efficient online inference, requiring only \textbf{5.35s} per minute of video, with \textbf{19.54s} per minute for memory construction. Moreover, \modelname{} reduces deployment cost by relying on compact high-level textual memory, reaching only \textbf{17.8 GB} peak VRAM on an NVIDIA A100, lower than Qwen3-VL-8B and Video-RAG.

To provide more intuitive insights, we present several representative examples in Figure~\ref{fig:example}. 
For \textit{Temporal Reasoning}, \modelname{} primarily operates over knowledge graph in Symbolic Schema, where high level semantic abstractions and relational structure allow it to recover the temporal order of actions without directly revisiting low level visual details. 
For \textit{Action Recognition}, \modelname{} further drills down into Episodic Stream, where temporally localized visual evidence, aligned with event-level summaries, enables the recognition of more fine-grained actions in the current scene. 
For detail-sensitive tasks such as \textit{Details Counting}, \modelname{} descends all the way to Sensory Buffer, where concrete visual cues at a finer granularity can be retrieved to support precise verification. 

Overall, these examples illustrate that the pyramidal multimodal memory of \modelname{} supports a coarse-to-fine retrieval process across memory layers, allowing the model to progressively move from abstract semantic reasoning to detailed perceptual verification as task demands increase. This design not only improves prediction accuracy, but also effectively exploits the complementary strengths of textual and visual representations.

%% file: sections/conclusion.tex
\section{Conclusion}
\label{sec:conclusion}
In this work, we present \modelname{}, a pyramidal multimodal memory framework grounded in Fuzzy-Trace Theory. By structurally decoupling verbatim visual details from gist semantic schemas, \modelname{} effectively bridges the gap between fine-grained perception and high-level cognition. To govern memory construction, we propose SIB-GRPO, an information-theoretic approach for dynamic, redundancy-aware compression. Complementing this, we introduce an entropy-driven top-down retrieval strategy that adaptively drills down from abstract symbolic schemas to fine-grained sensory details under high uncertainty, ensuring efficient and precise information access. Extensive experiments demonstrate that \modelname{} achieves state-of-the-art performance and robust generalization, inspiring foundational cognitive infrastructure for long-horizon autonomous agents.

\section*{Limitations}
\label{sec:limitations}
While \modelname{} demonstrates robust performance in long-horizon video understanding, we identify several avenues for future optimization and research. (i) \emph{Computational Overhead vs. Reasoning Depth:} Our hierarchical architecture prioritizes precise, multi-granularity reasoning, which naturally incurs a higher computational cost during the construction phase compared to flat, compression-heavy models. However, the modular nature of the Sensory, Episodic, and Symbolic layers allows for asynchronous processing and parallelization. Future work will explore distilling the memory construction pipeline to further reduce latency for resource-constrained edge deployment. (ii) \emph{Dependency on Upstream Perception:} As a modular system, \modelname{} benefits from the rapid advancements in upstream vision encoders and captioners. While the system effectively filters irrelevant information via top-down retrieval, we anticipate that integrating stronger, end-to-end trained perception backbones will further enhance the system's robustness against visual artifacts. (iii) \emph{Generalization to Unsupervised Scenarios:} The current memory manager utilizes task-driven reinforcement learning (SIB-GRPO) to align memory retention with reasoning needs. Extending this mechanism to fully unsupervised or self-supervised settings—where explicit task signals are absent—represents an exciting direction for enabling autonomous, ``lifelong'' learning agents. (iv) \emph{Evolution towards Lifelong Agents:} Our evaluation focuses on standard long-video benchmarks. To better address true real-world deployment involving continuous, multi-session interactions with distribution shifts, we plan to extend \modelname{} to support dynamic memory updating and forgetting mechanisms better suited for open-ended, continuous agentic scenarios.

\section*{Ethics Statement}
Our research advances the capability of multimodal agents to process and remember long-form video content. We recognize the importance of responsible AI development and address the following considerations. (i) \emph{Privacy and Data Protection:} Memory-augmented agents inevitably process visual data that may contain personally identifiable information (PII). While our experiments utilize public, consented datasets, real-world deployment requires strict adherence to data minimization principles. We advocate for implementing local storage solutions and rigorous access controls to prevent unintended data disclosure. (ii) \emph{Bias Mitigation:} Intelligent agents may inherit biases present in the training data or upstream foundation models. The selective nature of memory construction could theoretically retain biased evidence if not carefully managed. We encourage continuous monitoring of the memory selection policy and the adoption of diverse benchmarks to ensure fair and representative reasoning outcomes. (iii) \emph{Responsible Deployment:} As with any powerful multimodal system, there is a potential for misuse in high-stakes decision-making. We emphasize that \modelname{} is designed as an assistive tool to augment human capabilities. Deployments in sensitive domains should always incorporate human-in-the-loop oversight to ensure reliability and accountability.

\section*{Use of AI Assistants}
We used large language models (e.g., ChatGPT and Gemini) in a lawful and policy-compliant manner solely for non-substantive assistance such as translation and language polishing of the manuscript. They were not used to generate experimental results, derive scientific claims, or make methodological decisions; all technical content and conclusions are authored and verified by the authors.

\section*{Acknowledgments}
We thank the anonymous reviewers and chairs for their efforts and constructive suggestions. 
This work is supported in part by the National Natural Science Foundation of China under grants 62521006, 624B2088, 62536003, 62571298, and 62576122.

%% file: appendix_sections/appendix.tex
\appendix

\section{Proof of the Variational IB Bounds}
\label{app:proof}

We consider a stochastic encoder (memory manager) $p_{\theta}(m\mid x)$
that maps sensory input $X$ to an episodic representation $M$.
We assume the standard IB Markov structure
$Y \leftrightarrow X \leftrightarrow M$, hence
\begin{align}
p_{\theta}(x,y,m)
&= p(x,y)\,p_{\theta}(m\mid x), \\
p_{\theta}(y\mid x,m)
&= p(y\mid x).
\end{align}
The induced marginals are
\begin{align}
p_{\theta}(m)
&= \int p(x)\,p_{\theta}(m\mid x)\,dx, \\
p_{\theta}(m,y)
&= \int p(x,y)\,p_{\theta}(m\mid x)\,dx.
\end{align}

We introduce (i) a variational decoder $q_{\phi}(y\mid m)$
to approximate $p_{\theta}(y\mid m)$, and
(ii) a variational prior $r(m)$
to approximate the intractable marginal $p_{\theta}(m)$.
Define
\vspace{-0.8em}

\begin{align}
\mathcal{L}_{\text{p}}(\theta,\phi)
& \triangleq
\mathbb{E}_{p(x,y)\,p_{\theta}(m\mid x)}
\!\left[\log q_{\phi}(y\mid m)\right], \\
\mathcal{L}_{\text{c}}(\theta)
& \triangleq
\mathbb{E}_{p(x)}
\!\left[
D_{\mathrm{KL}}
\!\left(p_{\theta}(m\mid x)\,\|\,r(m)\right)
\right].
\end{align}

\subsection{Lower bound on $I(M;Y)$}

\paragraph{Step 1: rewrite mutual information}
\begin{align}
I(M;Y)
&= \mathbb{E}_{p_{\theta}(m,y)}
   \!\left[
   \log \frac{p_{\theta}(y\mid m)}{p(y)}
   \right] \nonumber\\
&= \mathbb{E}_{p_{\theta}(m,y)}
   \!\left[\log p_{\theta}(y\mid m)\right]
   + H(Y).
\label{eq:imy_decomp}
\end{align}

\paragraph{Step 2: variational lower bound via KL non-negativity}
For each $m$, by the non-negativity of KL divergence,
\begin{equation}
\hspace*{-0.4em}%
\begin{aligned}
&\hphantom{=}\, D_{\mathrm{KL}}
\!\left(
p_{\theta}(y\mid m)\,\|\,q_{\phi}(y\mid m)
\right) \\
&= \mathbb{E}_{p_{\theta}(y\mid m)}
\!\left[
\log p_{\theta}(y\mid m)
-
\log q_{\phi}(y\mid m)
\right] \\
&\ge 0.
\end{aligned}
\label{eq:kl_nonneg}
\end{equation}
which implies

\vspace{-1em}

\begin{equation}
\mathbb{E}_{p_{\theta}(m,y)}
\!\left[\log p_{\theta}(y\mid m)\right]
\ge
\mathbb{E}_{p_{\theta}(m,y)}
\!\left[\log q_{\phi}(y\mid m)\right]
\label{eq:imy_var}
\end{equation}

\paragraph{Step 3: match the training sampling form}
Using \(p_{\theta}(m,y)=\int p(x,y)\,p_{\theta}(m\mid x)\,dx\), we have
\begin{align}
\mathbb{E}_{p_{\theta}(m,y)}
   \!\left[\log q_{\phi}(y\mid m)\right]
&= \mathbb{E}_{p(x,y)\,p_{\theta}(m\mid x)}
   \!\left[\log q_{\phi}(y\mid m)\right] \nonumber\\
&= \mathcal{L}_{\text{p}}(\theta,\phi).
\label{eq:lp_equiv}
\end{align}

Combining with \eqref{eq:imy_decomp}--\eqref{eq:imy_var},
\begin{align}
I(M;Y)
\ge
\mathcal{L}_{\text{p}}(\theta,\phi) + H(Y).
\end{align}

\subsection{Upper bound on $I(X;M)$}

\paragraph{Step 1: rewrite mutual information}
\begin{align}
I(X;M)=
\mathbb{E}_{p(x)}
\!\left[
D_{\mathrm{KL}}
\!\left(
p_{\theta}(m\mid x)\,\|\,p_{\theta}(m)
\right)
\right].
\label{eq:ixm_kl}
\end{align}

\paragraph{Step 2: relate $\mathcal{L}_{\text{c}}(\theta)$ and $I(X;M)$}
\begin{align}
\mathcal{L}_{\text{c}}(\theta) 
&=
\mathbb{E}_{p(x,m)}
\!\left[
\log \frac{p_{\theta}(m\mid x)}{p_{\theta}(m)}
\right]
+
\mathbb{E}_{p(x,m)}
\!\left[
\log \frac{p_{\theta}(m)}{r(m)}
\right] \nonumber\\
&=
I(X;M)
+
\mathbb{E}_{p_{\theta}(m)}
\!\left[
\log \frac{p_{\theta}(m)}{r(m)}
\right] \nonumber\\
&=
I(X;M)
+
D_{\mathrm{KL}}
\!\left(
p_{\theta}(m)\,\|\,r(m)
\right) \nonumber\\
&\ge I(X;M).
\label{eq:lc_ge_ixm}
\end{align}

Therefore,
\begin{align}
I(X;M)\le \mathcal{L}_{\text{c}}(\theta).
\end{align}

\subsection{Variational objective}

Combining the above bounds,
\begin{align}
&\hphantom{\le}\, I(X;M)-\beta I(M;Y) \nonumber\\
&\le
\mathcal{L}_{\text{c}}(\theta)
-
\beta\Big(\mathcal{L}_{\text{p}}(\theta,\phi)+H(Y)\Big).
\label{eq:ib_upper}
\end{align}

Since $H(Y)$ is constant w.r.t. $(\theta,\phi)$,
minimizing $I(X;M)-\beta I(M;Y)$ is equivalent to maximizing
\begin{align}
\max_{\theta,\phi}\quad
\beta\,\mathcal{L}_{\text{p}}(\theta,\phi)
-
\mathcal{L}_{\text{c}}(\theta).
\end{align}
This completes the proof.

\paragraph{Remark: Action-output correspondence}
\label{abb:remark}
To avoid ambiguity, we clarify that the operator output $a_{t,i}\in\mathcal{O}$ uniquely determines
the update outcome of the episodic stream. Thus, for a fixed sensory buffer $\mathcal{M}_{\mathrm{sens}}$
and fixed update rules, specifying the action sequence is sufficient to determine the resulting episodic memory.

Concretely, for each sensory item $m_{t,i}$ and the current latest node $e^\star$, the decision operator
$\psi(\cdot)$ produces an action $a_{t,i}$, and the stream is updated by a deterministic transition
\begin{equation}
\begin{aligned}
e^{\star}_{t,i}
&=
T\!\left(e^{\star}_{t,i-1},\, m_{t,i},\, a_{t,i}\right),\\
\mathcal{M}_{\mathrm{epi}}^{t,i}
&=
U\!\left(\mathcal{M}_{\mathrm{epi}}^{t,i-1},\, m_{t,i},\, a_{t,i}\right),
\end{aligned}
\end{equation}
where $T(\cdot)$ and $U(\cdot)$ are deterministic functions parameterized by the selected action
$a_{t,i}\in\{\textsc{ADD\_NEW},\textsc{MERGE},\textsc{DISCARD}\}$.
Therefore, the final episodic stream is a deterministic function of the action sequence:
\begin{equation}
\mathcal{M}_{\mathrm{epi}}
=
F\!\left(\mathcal{M}_{\mathrm{sens}},\, \{a_{t,i}\}\right).
\end{equation}
In other words, under fixed rules \((T,U)\), the action space \(\mathcal{O}\) and the resulting episodic memory
\(\mathcal{M}_{\mathrm{epi}}\) provide equivalent descriptions of the same compression process.

\section{Dataset and Benchmark Details}
\label{app:dataset}

\begin{figure*}[t]
    \centering
    \includegraphics[width=\linewidth]{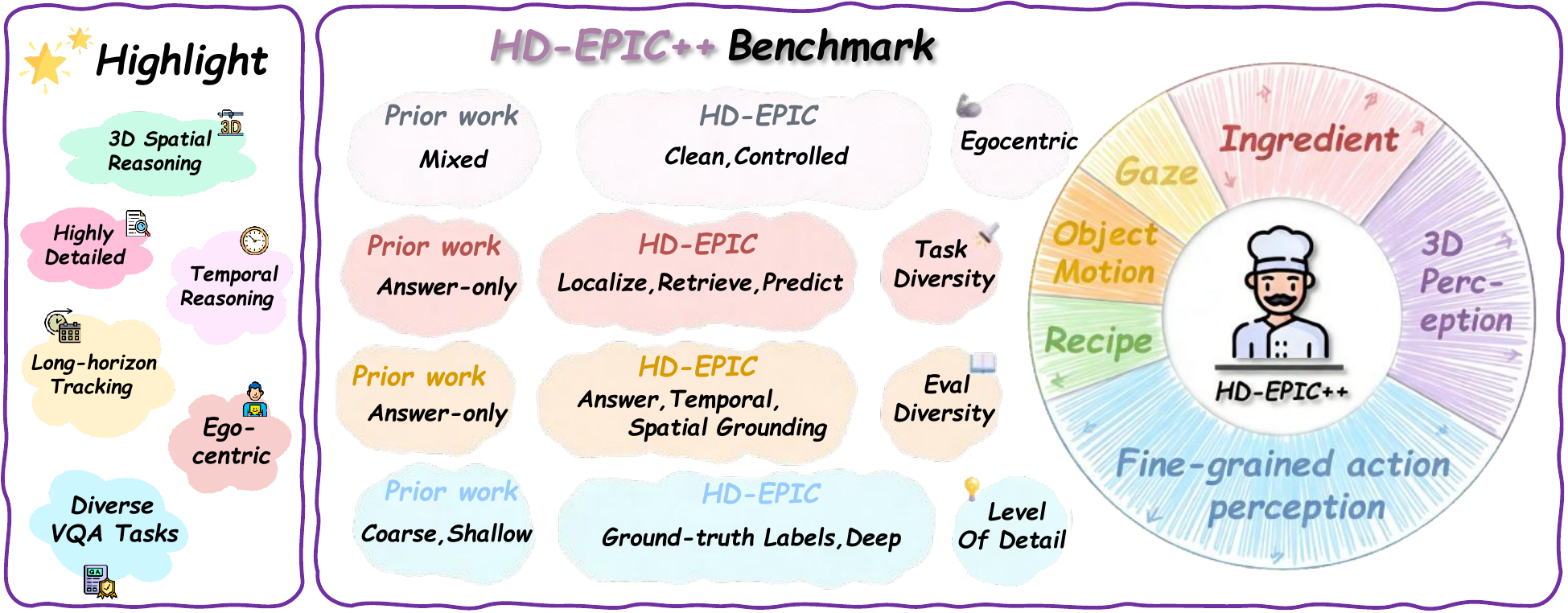}
    \caption{\textbf{\textsc{HD-EPIC++}} is an egocentric long-horizon kitchen video benchmark with highly detailed annotations, covering fine-grained action perception, temporal reasoning, 3D spatial understanding, object motion, gaze, and diverse VQA tasks (e.g., recipes and ingredients).}
    \label{fig:hd-epic++}
\end{figure*}

\subsection{Proposed HD-EPIC++}\label{app:hdepicpp}

\paragraph{Motivation and split protocol}
Our method requires \textsc{SIB-GRPO} finetuning, which necessitates a dedicated training split. In addition, to enable straightforward and reproducible evaluation, we adopt a fixed train/test partition so that results can be directly compared without re-splitting. Concretely, we re-organize the original \textsc{HD-EPIC} videos into \textbf{105} training videos and \textbf{51} test videos, totaling \textbf{156} videos, and use the training split for \textsc{SIB-GRPO} finetuning while reporting all benchmark performance on the held-out test split.

\paragraph{Dataset overview}
We introduce \textsc{HD-EPIC++}, an improved dataset built upon \textsc{HD-EPIC}, extending it with denser, fine-grained annotations tailored for long-horizon, procedure-centric video understanding. Compared to the original \textsc{HD-EPIC}, \textsc{HD-EPIC++} emphasizes richer supervision signals that support not only recognition and temporal grounding, but also structured reasoning over procedures, entities, and interactions.

\paragraph{Dense annotations}
\textsc{HD-EPIC++} provides dense supervision covering \textbf{7} key annotation types: \textit{(i) Recipe} (identify, retrieve, and localize recipes and steps), \textit{(ii) Ingredient} (track ingredient usage, weight, timing, and order), \textit{(iii) Nutrition} (analyze ingredient nutrition and its evolution throughout recipes), \textit{(iv) Fine-Grained Action} (understand the what/how/why of actions), \textit{(v) 3D Perception} (reason about object positions in 3D space), \textit{(vi) Object Motion} (track object movements across long video sequences), and \textit{(vii) Gaze} (estimate fixation points and anticipate future interactions).

\paragraph{VQA benchmark construction}
Leveraging the dense annotations, we construct a multiple-choice VQA benchmark to evaluate long-context, multimodal reasoning. For each question type, we use a \textbf{5-way} multiple-choice format. We design \textbf{30} question prototypes, which instantiate into \textbf{26,650} multiple-choice questions. To increase difficulty and reduce shortcut learning, we sample \textbf{hard negatives} from within the dataset based on the underlying annotations.

\paragraph{Scalability and intended impact}
The benchmark is designed to be large-scale yet tractable for evaluation with closed-source VLMs. Due to the annotation density, we estimate an upper bound of approximately \textbf{100,000} unique questions that can be generated via additional instantiations. We expect \textsc{HD-EPIC++} to facilitate systematic evaluation of (i) long-horizon procedural understanding, (ii) entity/state tracking over time, and (iii) grounded multimodal reasoning under realistic distractors.

\section{Salient Key Sub-clip Extraction for Sensory Buffer}
\label{app:sensory-keyframe}

For each clip $c_t$ with frames $\{f_{t,i}\}_{i=1}^{|c_t|}$, we quantify inter-frame variation by
\begin{equation}
d_{t,i}=\frac{1}{|\Omega|}\sum_{p\in\Omega}\left\|f_{t,i}(p)-f_{t,i-1}(p)\right\|_1,
\end{equation}
where $\Omega$ denotes the pixel grid. We then compute the clip-level statistics
\begin{equation}
\mu_t=\mathbb{E}_i[d_{t,i}],\qquad \sigma_t=\mathrm{Std}_i[d_{t,i}],
\end{equation}
and select salient indices via a simple adaptive threshold:
\begin{equation}
\mathcal{S}_t=\{\, i \mid d_{t,i}>\mu_t+\sigma_t \,\}.
\end{equation}

\paragraph{Key sub-clip construction}
For each $i\in\mathcal{S}_t$, we extract a short temporal window centered at $i$ as a key sub-clip (sensory evidence), and record its temporal location $\tau_{t,i}$ (e.g., the timestamp of the center frame).

\paragraph{Near-duplicate suppression}
To avoid redundant evidence while preserving salient dynamics, we suppress near-duplicate candidates among $\mathcal{S}_t$ by enforcing a minimum temporal separation. Concretely, let $\mathcal{S}_t$ be sorted by decreasing $d_{t,i}$, and iteratively keep an index $i$ only if it is at least $\Delta$ frames away from all previously kept indices; otherwise it is discarded. This yields a compact set of key sub-clips that covers salient changes without excessive overlap.

\paragraph{Memory tuple instantiation}
Each key sub-clip centered at $i$ is encoded into a visual representation $\mathbf{v}_{t,i}$ (e.g., by a video encoder), paired with a text trace $\mathbf{l}_{t,i}$ (aligned subtitles or automatic captioning), and stored with its temporal location $\tau_{t,i}$.

\section{Implementation Details}
\label{abb:imp}

\begin{table}[t]
\centering
\begin{tabular}{ll}
\noalign{\hrule height 1.2pt}

\rowcolor{green!10}
\multicolumn{2}{c}{\textcolor{ForestGreen}{LoRA}}             \\ \hline
\rowcolor{green!4}
lora\_rank                & 64       \\
\rowcolor{green!4}
lora\_alpha               & 128      \\
\rowcolor{green!4}
lora\_dropout             & 0.05     \\ \hline \hline
\rowcolor{blue!10}
\multicolumn{2}{c}{\textcolor{RoyalBlue}{SIB-GRPO}}  \\ \hline
\rowcolor{blue!4}
epoch                     & 3        \\
\rowcolor{blue!4}
batch\_size               & 8        \\
\rowcolor{blue!4}
learning\_rate            & 1e-5 \\
\rowcolor{blue!4}
beta                      & 0.1      \\
\rowcolor{blue!4}
ppo\_clip\_epsilon        & 0.2      \\
\rowcolor{blue!4}
use\_importance\_sampling & TRUE     \\
\rowcolor{blue!4}
kl\_penalty\_coef         & 0.1      \\
\rowcolor{blue!4}
save\_steps               & 100      \\ \hline\hline
\rowcolor{red!10}
\multicolumn{2}{c}{\textcolor{BrickRed}{Reasoning}} \\ \hline
\rowcolor{red!4}
top\_k\_sym               & 5        \\
\rowcolor{red!4}
top\_k\_epi               & 2        \\
\rowcolor{red!4}
top\_k\_sen               & 1        \\
\rowcolor{red!4}
$\gamma$                       & 0.72     \\ \hline \hline
\end{tabular}
\caption{Training and inference hyperparameters.}
\label{tab:hyper}
\end{table}

\paragraph{Hyperparameter Settings}
\label{abb:hyper}
We fine-tune the model with SIB-GRPO on the training split of HD-EPIC++, and perform LoRA adaptation within the SWIFT framework. The hyperparameters used for training and inference are summarized in Table~\ref{tab:hyper}.

\paragraph{Evaluation protocol}
\label{abb:eval}
The same evaluation protocol as Flash-VStream \cite{VStream-QA} is followed. Since VStream-QA consists of open-ended questions, GPT-4o-mini (Hurst et al., 2024) is adopted as an automatic judge. Given a model prediction, whether the prediction is correct is judged, and a score between 0 and 5 is assigned. Accordingly, two metrics are reported on VStream-QA: (i) Accuracy, which indicates whether the prediction is judged correct, and (ii) Score, which is computed as the average of the assigned scores. 

\begin{figure}[!t]
    \centering
    \includegraphics[width=\linewidth]{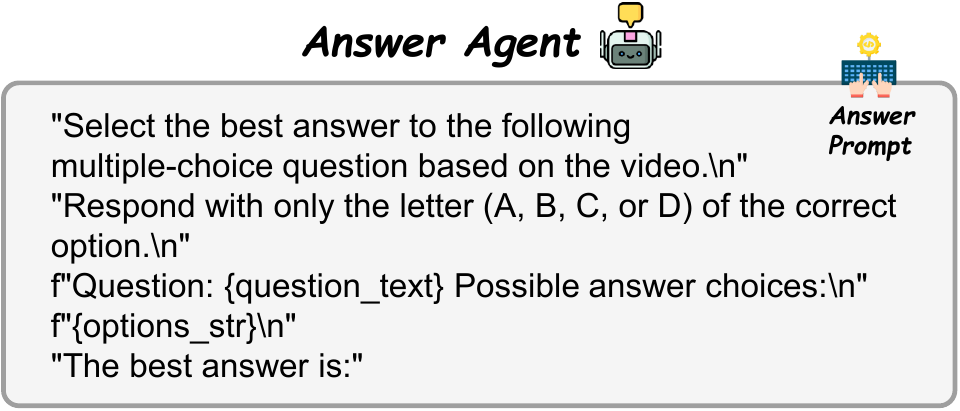}
    \caption{Answer Agent prompt template instructing the model to select the best option for a video-based multiple-choice question and respond only with the corresponding letter (A--D).}
    \label{fig:answer_prompt}
\end{figure}

\begin{figure}[!t]
    \centering
    
    \begin{subfigure}[t]{\linewidth}
        \centering
        \includegraphics[width=\linewidth]{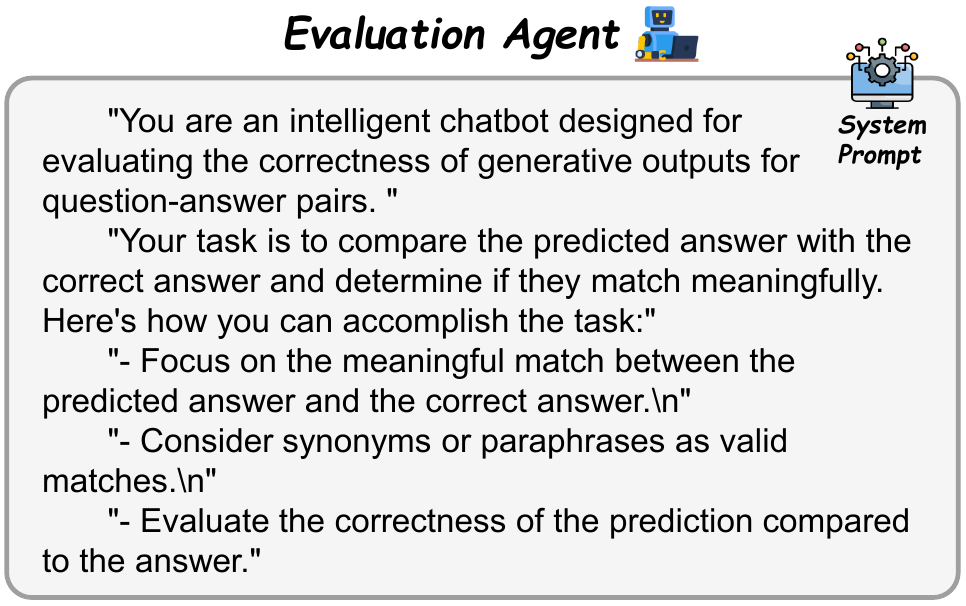}
        \caption{System prompt for VStream-QA evaluation agent, instructing the model to judge whether a generated answer matches the ground-truth answer (allowing paraphrases and synonyms) and to assess correctness based on meaningful semantic alignment.}
        \label{fig:vstream_sys_prompt}
    \end{subfigure}

    \vspace{2mm}

    \begin{subfigure}[t]{\linewidth}
        \centering
        \includegraphics[width=\linewidth]{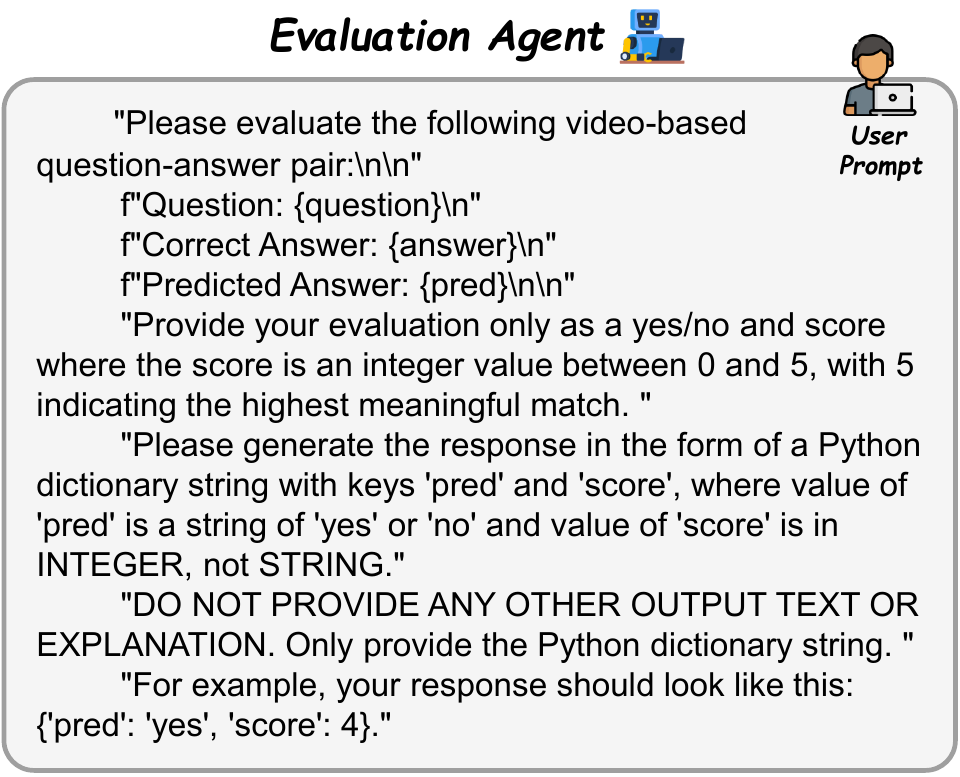}
        \caption{User prompt for VStream-QA evaluation agent, requesting a strict Python-dictionary output with a yes/no correctness label and an integer similarity score from 0 to 5 for a video-based question--answer pair.}
        \label{fig:vstream_user_prompt}
    \end{subfigure}

    \caption{Prompts used by VStream-QA evaluation agent (system and user prompts).}
    \label{fig:vstream_eval_prompts}
\end{figure}

\section{Details of Prompts}
\label{app:prompt}

\noindent\textbf{Prompt sources.} The Answer Agent prompt shown in Figure~\ref{fig:answer_prompt} is adapted from the prompt template described in the official Qwen3-VL technical report~\cite{qwen3vl}. We adopt this template to remain consistent with our choice of Qwen3-VL as the base model and to follow the recommended instruction and formatting conventions for multimodal question answering. 
Similarly, the evaluation agent prompts for VStream-QA~\cite{VStream-QA} (the system and user prompts in Figure~\ref{fig:vstream_sys_prompt} and Figure~\ref{fig:vstream_user_prompt}) are taken from prompts released on the official VStream-QA website. We use these prompts without altering their intended evaluation protocol, ensuring that our results are directly comparable to the standard VStream-QA evaluation setup.

\section{Additional Experiments}\label{abb:add_exp}

\begin{table}[t]
\centering
\resizebox{\linewidth}{!}{%
\begin{tabular}{llcccc}
\hline
\multirow{2}{*}{ID} & \multirow{2}{*}{Method} & \multicolumn{4}{c}{Video-MME} \\ \cline{3-6}
                   &                         & Short & Medium & Long & Overall \\ \hline
(I)   & w/o $ SIB-GRPO $      & 80.1  & 73.9   & 71.9 & 75.3    \\
(II)  & w/o $ Visual Memory $ & 76.5  & 70.1   & 66.1 & 70.9    \\
(III) & w/o $ Text Memory$   & 80.5  & 74.2   & 72.8 & 75.8    \\
(IV)  & w/o $ Symbolic$     & 81.2  & 74.6   & 73.5 & 76.4    \\
(V)   & w/o $ Episodic$      & 80.2  & 73.6   & 71.2 & 75.0    \\
(VI)  & w/o $ Sensor$        & 78.6  & 71.9   & 70.2 & 73.6    \\
(VII) & w/o $Memory$        & 79.2  & 72.6   & 70.9 & 74.2    \\
\rowcolor{blue!6}
(VIII)& $\textbf{\modelname{} (full)} $    & \bm{82.8$\pm$ 0.2}  & \bm{75.8$\pm$ 0.2}   & \bm{75.7$\pm$ 0.3} & \bm{78.1$\pm$ 0.2}    \\ \hline
\end{tabular}%
}
\caption{Ablation studies of individual modules on Video-MME with subtitles.}
\label{tab:abb-abl}
\end{table}

\subsection{Ablation Studies}
Table~\ref{tab:abb-abl} reports module-wise ablations on Video-MME under the subtitle setting.
The full model (\modelname{}) achieves \textbf{82.8/75.8/75.7} on Short/Medium/Long and \textbf{78.1} overall.
Removing any component consistently degrades performance, indicating that each module contributes positively.

\begin{itemize}[leftmargin=*]
  \item \textbf{Visual vs.\ Text Memory.}
  Visual Memory is the most critical component: removing it drops Overall from 78.1 to 70.9,
  with an even larger decrease on Long (75.7 $\rightarrow$ 66.1, \textbf{-9.6}).
  In contrast, removing Text Memory yields a smaller but consistent drop (Overall 78.1 $\rightarrow$ 75.8, \textbf{-2.3}),
  suggesting that subtitles provide useful high-level cues, yet fine-grained and verifiable visual evidence remains indispensable,
  especially for long videos.

  \item \textbf{Training / Memory Management (SIB-GRPO).}
  Removing SIB-GRPO reduces Overall to 75.3 (\textbf{-2.8}) and Long to 71.9 (\textbf{-3.8}),
  showing that the redundancy-aware memory optimization is particularly beneficial as temporal context grows.

  \item \textbf{Hierarchical Memory Components (Sensor / Episodic / Symbolic).}
  Among three memory layers, \emph{Sensor} and \emph{Episodic} have stronger impact on longer videos:
  removing Sensor yields Overall 73.6 (\textbf{-4.5}) and Long 70.2 (\textbf{-5.5}),
  while removing Episodic gives Overall 75.0 (\textbf{-3.1}) and Long 71.2 (\textbf{-4.5}).
  Removing Symbolic causes a smaller average drop (Overall 76.4, \textbf{-1.7}), but remains consistently helpful,
  likely supporting higher-level temporal/relational reasoning.

  \item \textbf{Disabling memory entirely.}
  Without the memory mechanism, performance drops to Overall 74.2 (\textbf{-3.9}) and Long 70.9 (\textbf{-4.8}),
  indicating that explicit memory retrieval/organization improves robustness even when subtitles are available.
\end{itemize}

\begin{table}[t]
\centering
{
\begin{tabular}{lllll}
\hline
\multirow{2}{*}{Method} & \multicolumn{2}{c}{VS-Ego} & \multicolumn{2}{c}{VS-Movie} \\ \cline{2-5} 
                        & Acc.         & Sco.        & Acc.          & Sco.         \\ \hline
Video-ChatGPT           & 51.7         & 3.7         & 54.4          & 3.4          \\
MovieChat               & 52.2         & 3.4         & 39.1          & 2.3          \\
Chat-UniVi              & 50.9         & 3.8         & 54            & 3.4          \\
LLaMA-VID               & 54.8         & 3.9         & 51.4          & 3.4          \\
Flash-VStream           & 59           & 3.9         & 56.1          & 3.4          \\
\rowcolor{blue!6}
\modelname{}                  & 62.5         & 4.1         & 52.1          & 3.2          \\ \hline
\end{tabular}
}
\caption{Performance comparison on the VStream-QA benchmark, evaluated on the VS-Ego and VS-Movie splits in terms of accuracy (Acc.) and score (Sco.).}
\label{tab:vstreamqa}
\end{table}

\subsection{Descriptive Statistics}
For the main ablation in Table~\ref{tab:abb-abl}, we report mean accuracy with uncertainty estimated from repeated evaluations, shown as $\pm$ values. The full \modelname{} achieves \bm{82.8$\pm$0.2}, \bm{75.8$\pm$0.2}, and \bm{75.7$\pm$0.3} accuracy on the Short/Medium/Long subsets, respectively, with an overall score of \bm{78.1$\pm$0.2}. The relatively small variances across subsets indicate stable performance under the same evaluation protocol and suggest that the improvements of the full system over ablated variants are consistent rather than driven by outlier runs.

\subsection{Results on \textsc{VSteam-QA}}
Table~\ref{tab:vstreamqa} summarizes results on the VStream-QA benchmark, covering two splits: VS-Ego and VS-Movie.
Overall, \modelname{} achieves the best performance on the VS-Ego split, while its performance on VS-Movie is comparatively weaker, suggesting different challenges across domains.

\begin{itemize}[leftmargin=*]
  \item \textbf{VS-Ego (streaming egocentric videos).}
  \modelname{} attains the highest \textbf{Acc.} of \textbf{62.5} and the highest \textbf{Sco.} of \textbf{4.1}.
  Compared with the strongest baseline Flash-VStream (59.0 Acc., 3.9 Sco.), \modelname{} improves accuracy by \textbf{+3.5} and score by \textbf{+0.2}.
  This indicates that \modelname{} is particularly effective for streaming egocentric understanding, where long-horizon context accumulation and memory utilization are crucial.

  \item \textbf{VS-Movie (streaming movie clips).}
  On VS-Movie, \modelname{} reaches 52.1 Acc. and 3.2 Sco., \textbf{below} the best baselines (e.g., Video-ChatGPT/Chat-UniVi/Flash-VStream at $\sim$54.0--56.1 Acc. and 3.4 Sco.).
  In particular, it trails Flash-VStream by \textbf{-4.0} Acc. (52.1 vs.\ 56.1) and \textbf{-0.2} Sco.\ (3.2 vs.\ 3.4),
  suggesting that movie-style streaming QA relies on cues (e.g., scene cuts, dialogue patterns, narrative coherence) not fully captured by the current memory design.

  \item \textbf{Cross-split observation.}
  \modelname{} shows clear gains on VS-Ego, but a smaller advantage on VS-Movie,
  suggesting the memory-centric design generalizes better to egocentric streaming than to movie-style content under the current setup.

\end{itemize}

\begin{table}[t]
\centering
{
\begin{tabular}{l l}
\hline
\multicolumn{1}{c}{\multirow{2}{*}{\textbf{Method}}}
 & \multicolumn{1}{c}{\textbf{HD-EPIC++}} \\ \cline{2-2}
                                   & \multicolumn{1}{c}{\textbf{Accuracy}}  \\ \hline \hline
Qwen3-VL-8B \cite{qwen3vl}          & \multicolumn{1}{c}{25.88} \\
Qwen2.5-VL-7B  \cite{qwen3vl}      & \multicolumn{1}{c}{24.37} \\
LLaVA-Video-7B  \cite{qwen3vl}     & \multicolumn{1}{c}{25.37} \\
VideoLLaMA 3-7B \cite{Videollama3}  & \multicolumn{1}{c}{20.36} \\ 
Qwen3-VL-4B  \cite{qwen3vl}          & \multicolumn{1}{c}{24.91} \\
Qwen3-VL-2B  \cite{qwen3vl}         & \multicolumn{1}{c}{22.80}  \\
Qwen3-VL-8B (sft)         & \multicolumn{1}{c}{27.25}  \\
\rowcolor{purple!7}
\textbf{\modelname{} (Ours)}                              & \multicolumn{1}{c}{$\bm{30.28}$} \\ \hline \hline
\end{tabular}
}
\caption{Evaluation on the proposed \textbf{\textsc{HD-EPIC++}}.}
\label{tab:hd-epicpp}
\end{table}

\subsection{SFT vs. SIB-GRPO RL}
As shown in Table~\ref{tab:hd-epicpp}, supervised fine-tuning (SFT) provides a clear but limited improvement over the strong Qwen3-VL-8B baseline, increasing accuracy from 25.88 to 27.25 (+1.37). In contrast, our SIB-GRPO reinforcement learning delivers a substantially larger gain, achieving 30.28 (+3.03 over the same baseline, and +3.03 compared to the base model; +3.03 - 1.37 = +1.66 over SFT). This indicates that while SFT mainly helps the model better align with the training distribution and learn surface-level task patterns, SIB-GRPO further optimizes decision-making under the benchmark’s complex video understanding requirements, leading to more reliable long-horizon reasoning and higher-quality action grounding. Overall, the margin suggests that preference-driven reinforcement learning complements supervised learning by rewarding correct end-task behavior, making it more effective for HD-EPIC++ where errors often stem from compounding misinterpretations rather than insufficient visual feature extraction.